%% file: 0_main.tex
\documentclass{article}

\PassOptionsToPackage{numbers}{natbib}

\usepackage[final]{neurips_2022}

\usepackage[resetlabels]{multibib}
\newcites{sup}{Supplementary References}

\usepackage[utf8]{inputenc} 
\usepackage[T1]{fontenc}    
\usepackage{hyperref}       
\usepackage{url}            
\usepackage{booktabs}       
\usepackage{amsfonts}       
\usepackage{nicefrac}       
\usepackage{microtype}      
\usepackage{natbib}
\usepackage{amsmath, amsthm, amssymb} 
\usepackage{enumitem}
\usepackage[dvipsnames]{xcolor}
\usepackage[linesnumbered,ruled,vlined,noend]{algorithm2e}
\usepackage[noend]{algpseudocode}
\usepackage{pseudocode}
\usepackage{mathtools}
\usepackage{xspace}
\usepackage{multirow}
\usepackage{caption}
\usepackage{subcaption}
\usepackage{stmaryrd}
\usepackage{cryptocode}
\usepackage{hhline}
\usepackage{multicol}
\usepackage{sidecap}  
\sidecaptionvpos{figure}{c}
\usepackage{dblfloatfix}    
\usepackage{soul}
\usepackage{tikz}
\usepackage{wrapfig}

\title{LiteTransformerSearch: Training-free Neural Architecture Search for Efficient Language Models}
\vspace{-0.5cm}

\author{%
Mojan Javaheripi$^1$, Gustavo H. de Rosa$^2$, Subhabrata Mukherjee$^2$,\\
\textbf{ Shital Shah$^2$, Tomasz L. Religa$^3$, Caio C.T. Mendes$^2$,} 
\\ \vspace{0.1cm}
\textbf{Sebastien Bubeck$^2$, Farinaz Koushanfar$^1$,  Debadeepta Dey$^2$}
\\ \vspace{0.1cm}
  $^1$University of California San Diego, $^2$Microsoft Research, $^3$Microsoft \\
  \texttt{mojan@ucsd.edu, dedey@microsoft.com}
}

\def\mojan{\textcolor{magenta}}

\newcommand{\comment}[1]{}

\newcommand{\sys}[1]{{LTS}}

\SetCommentSty{mycommfont}
\newcommand*\circled[1]{\tikz[baseline=(char.base)]{
            \node[shape=circle,fill,inner sep=0.5pt] (char) {\textcolor{white}{#1}};}}
            
\newcommand{\jacobcov}{\texttt{jacob\_cov} }
\newcommand{\snip}{\texttt{snip} }
\newcommand{\grasp}{\texttt{grasp} }
\newcommand{\synflow}[1][]{\texttt{synflow#1} }

\newcommand{\fisher}{\texttt{fisher} }

\newcommand{\gradnorm}{\texttt{grad\_norm} }
\newcommand{\relulogdet}{\texttt{relu\_log\_det} }

\begin{document}

\maketitle

\vspace{-0.5cm}
\begin{abstract}
\vspace{-0.2cm}
The Transformer architecture is ubiquitously used as the building block of
large-scale autoregressive language models. However, finding
architectures with the optimal trade-off between task performance (perplexity) and hardware constraints like peak memory utilization and latency is non-trivial. This is exacerbated by the proliferation of various hardware. We leverage the somewhat surprising empirical observation that the number of decoder parameters in autoregressive Transformers has a high rank correlation with task performance, irrespective of the architecture topology. This observation organically induces a simple Neural Architecture Search (NAS) algorithm that uses decoder parameters as a proxy for perplexity without need for any model training. The search phase of our training-free algorithm, dubbed Lightweight Transformer Search (\sys{})\footnote{code available at \url{https://github.com/microsoft/archai/tree/neurips_lts/archai/nlp}}, can be run directly on target devices since it does not require GPUs. Using on-target-device measurements, \sys{} extracts the Pareto-frontier of perplexity versus any hardware performance cost.
We evaluate \sys{} on 
diverse devices from ARM CPUs to NVIDIA GPUs and two popular autoregressive Transformer backbones: GPT-2 and Transformer-XL. Results show that the perplexity of $16$-layer GPT-2 and Transformer-XL can be achieved with up to $1.5\times,2.5\times$ faster runtime and $1.2\times,2.0\times$ lower peak memory utilization. When evaluated in zero and one-shot settings, \sys{} Pareto-frontier models achieve higher average accuracy compared to the $350$M parameter OPT across $14$ tasks, with up to $1.6\times$ lower latency.
\sys{} extracts the Pareto-frontier in under $3$ hours while running on a commodity laptop. We effectively remove the carbon footprint of hundreds of GPU hours of training during search, offering a strong simple baseline for future NAS methods in autoregressive language modeling.
\end{abstract}

\vspace{-0.5cm}
\input{1_intro}

\input{2_related_work}

\input{3_methodology}
\input{4_experiments}

\input{5_conclusion}

\bibliographystyle{plainnat}
{
\small
\bibliography{ref}
}

\section*{Checklist}

\begin{enumerate}

\item For all authors...
\begin{enumerate}
  \item Do the main claims made in the abstract and introduction accurately reflect the paper's contributions and scope?
    \answerYes
  \item Did you describe the limitations of your work?
    \answerYes See Sections~\ref{sec:topology},~\ref{sec:conclusion}, and Appendix~\ref{sec:appdx_scaling}
  \item Did you discuss any potential negative societal impacts of your work?
    \answerYes See Appendix~\ref{sec:ethics}
  \item Have you read the ethics review guidelines and ensured that your paper conforms to them?
    \answerYes
\end{enumerate}

\item If you are including theoretical results...
\begin{enumerate}
  \item Did you state the full set of assumptions of all theoretical results?
    \answerNA{}
        \item Did you include complete proofs of all theoretical results?
    \answerNA{}
\end{enumerate}

\item If you ran experiments...
\begin{enumerate}
  \item Did you include the code, data, and instructions needed to reproduce the main experimental results (either in the supplemental material or as a URL)?
    \answerYes The source code for \sys{} is available at \url{https://github.com/microsoft/archai/tree/neurips_lts/archai/nlp}
  \item Did you specify all the training details (e.g., data splits, hyperparameters, how they were chosen)?
    \answerYes See Appendix~\ref{sec:appdx_training}
        \item Did you report error bars (e.g., with respect to the random seed after running experiments multiple times)?
    \answerNA{}
        \item Did you include the total amount of compute and the type of resources used (e.g., type of GPUs, internal cluster, or cloud provider)?
    \answerYes
\end{enumerate}

\item If you are using existing assets (e.g., code, data, models) or curating/releasing new assets...
\begin{enumerate}
  \item If your work uses existing assets, did you cite the creators?
    \answerYes{} See Appendix~\ref{sec:appdx_training}
  \item Did you mention the license of the assets?
    \answerNA{} All used code are open-source and publicly available
  \item Did you include any new assets either in the supplemental material or as a URL?
    \answerYes
  \item Did you discuss whether and how consent was obtained from people whose data you're using/curating?
    \answerNA{}
  \item Did you discuss whether the data you are using/curating contains personally identifiable information or offensive content?
    \answerNA{}
\end{enumerate}

\item If you used crowdsourcing or conducted research with human subjects...
\begin{enumerate}
  \item Did you include the full text of instructions given to participants and screenshots, if applicable?
    \answerNA{}
  \item Did you describe any potential participant risks, with links to Institutional Review Board (IRB) approvals, if applicable?
    \answerNA{}
  \item Did you include the estimated hourly wage paid to participants and the total amount spent on participant compensation?
    \answerNA{}
\end{enumerate}

\end{enumerate}

\newpage


\appendix
\input{appendix}

\end{document}

%% file: 1_intro.tex
\vspace{-0.1cm}
\section{Introduction}\label{sec:intro}
\vspace{-0.1cm}
The Transformer architecture~\citep{vaswani2017attention} has been used as the de-facto building block of most pre-trained language models like GPT~\citep{brown2020language}. A common problem arises when one tries 
to create smaller versions of Transformer models for edge or real-time applications (e.g. text prediction) with strict memory and latency constraints: it is not clear
what the architectural hyperparameters should be, e.g., number of attention heads, number of layers, embedding dimension, and the inner dimension of the feed forward network, etc. This problem is exacerbated if each
Transformer layer is allowed the freedom to have different values for these settings. 
This results in a combinatorial explosion of architectural hyperparameter choices and a large heterogeneous search space. For instance, the
search space considered in this paper consists of over $10^{54}$ possible architectures.

Neural Architecture Search (NAS) is an organic solution due to its ability to automatically search
through candidate models with multiple conflicting objectives like latency vs. task performance.
The central challenge in NAS is the prohibitively expensive function evaluation, i.e., evaluating each architecture requires training it on the dataset at hand. Thus it is often infeasible to evaluate more than a handful of architectures during the search phase. 
Supernets~\cite{pham2018efficient} have emerged as a dominant 
paradigm in NAS which combine all 
possible architectures into a single graph and jointly train them using weight-sharing. 
Nevertheless, supernet training imposes constraints
on the expressiveness of the search space~\cite{ning2021evaluating} and is often memory-hungry~\cite{pcdarts,ofa,analyzingnas}
as it creates large networks during search. Additionally, training supernets is non-trivial as children architectures may interfere with each other and the ranking between sub-architectures based on task performance is not preserved~\cite{ning2021evaluating}\footnote{See \citep{ning2021evaluating} for a comprehensive treatment of the difficulties of training supernets.}. 

We consider a different approach by proposing a training-free proxy that provides a highly accurate ranking of candidate architectures during NAS without need for costly function evaluation or supernets. Our scope is NAS for efficient autoregressive Transformers used in language modeling. We design a lightweight search method that is target hardware-aware and outputs a gallery of 
models on the Pareto-frontier of perplexity versus hardware metrics. We term this method Lightweight Transformer Search (\sys{}). \sys{} relies on our somewhat surprising observation: {\textit{the decoder parameter count
has a high rank correlation with the perplexity of fully trained autoregressive Transformers.}} 

Given a set of autoregressive Transformers, one can accurately rank them using decoder parameter count as the proxy for perplexity. Our observations are also well-aligned with the power laws in~\cite{kaplan2020scaling}, shown for homogeneous autoregressive Transformers, i.e., when all decoder layers have the same configuration. We provide extensive experiments that establish a high rank correlation between perplexity and decoder parameter count for \emph{both} homogeneous and heterogeneous search spaces. 


\begin{SCfigure}[50][t]
    \centering
    \includegraphics[width=0.45\columnwidth]{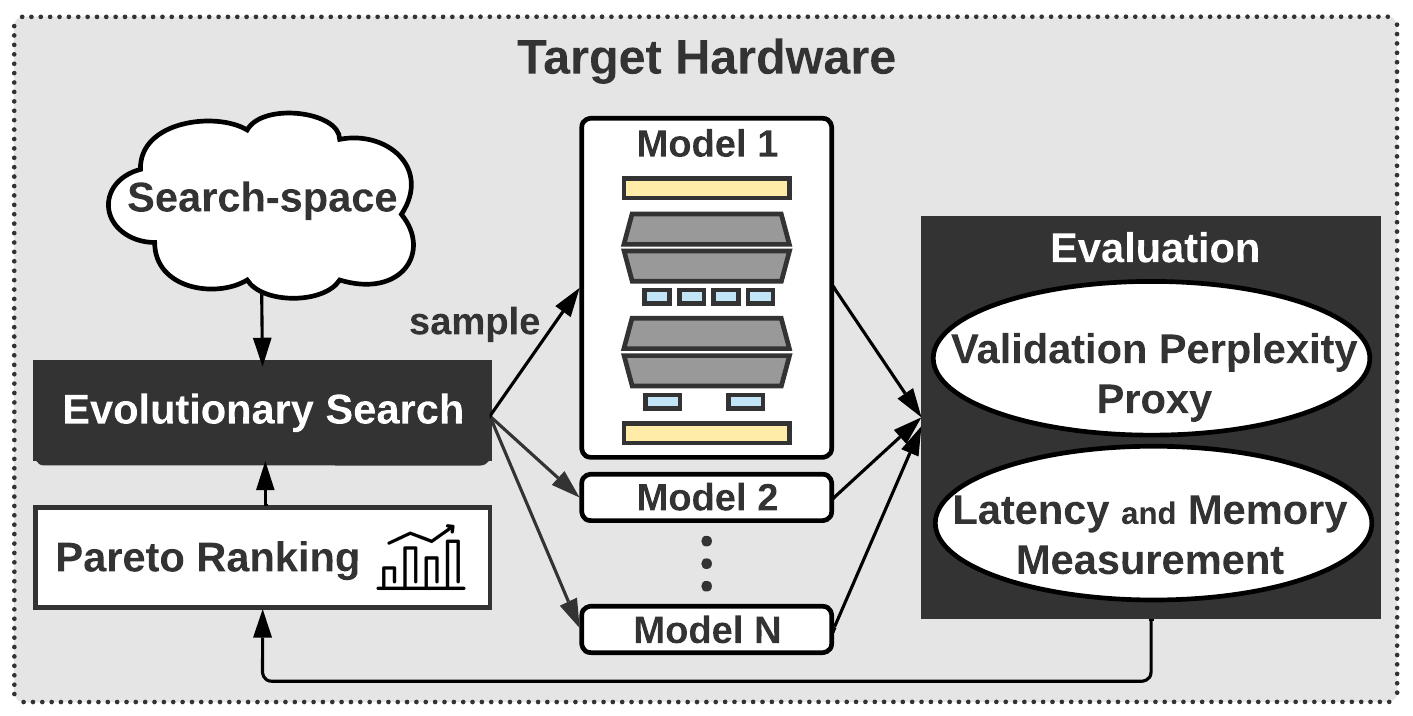}
    \caption{High-level overview of \sys{}. We propose a training-free zero-cost proxy for evaluating the validation perplexity of candidate architectures. Pareto-frontier search is powered by evolutionary algorithms which use the proposed proxy along with real latency and memory measurements on the target hardware to evaluate sampled architectures.}
    \label{fig:overview}
\vspace{-0.5cm}
\end{SCfigure}

The above phenomenon coupled with the fact that a candidate architecture's hardware performance can be measured on the target device 
leads to a training-free search procedure: {\textit{pick one's favorite discrete search algorithm 
(e.g. evolutionary search), sample candidate architectures from the search space; count their decoder parameters as a proxy for task performance (i.e., perplexity); measure their hardware performance (e.g., latency and memory) directly on the target device; and progressively create a Pareto-frontier estimate.}} While we have chosen a reasonable search algorithm in this work, one can plug and play any Pareto-frontier search method such as those in \cite{guerrero2021bagofbaselines}.

Building upon these insights, Figure~\ref{fig:overview} shows a high-level overview of \sys{}. We design the first training-free Transformer search that is performed entirely on the target (constrained) platform. 
As such, \sys{} easily performs a multi-objective NAS where several underlying hardware performance metrics, e.g., latency and peak memory utilization, are simultaneously optimized. Using our training-free proxy, we extract the $3$-dimensional Pareto-frontier of perplexity versus latency and memory in a record-breaking time of $<3$ hours on a commodity Intel Core i7 CPU. Notably, \sys{} eliminates the carbon footprint from hundreds of GPU hours of training associated with legacy NAS methods. 

\comment{
Once the Pareto-frontier models are identified, the user can pick a model based on their desired hardware constraints and fully train it on the target dataset. We note that the cost of training this model is lower than the cost of training large supernets, particularly since our NAS is focused on low-latency and memory-efficient Transformers. \sys{} also avoids the many difficulties associated with training supernets
\footnote{See \citep{ning2021evaluating} for a comprehensive treatment of the difficulties of training supernets.}.
}

To corroborate the effectiveness of our proxy, we train over $2900$ Transformers on three large language modeling benchmark datasets, namely, WikiText-103~\cite{merity2016pointer}, One Billion Word~\cite{chelba2013one}, and Pile~\cite{gao2020pile}. We use \sys{} to search for Pareto-optimal architectural hyperparameters in two popularly used autoregressive Transformer backbones, namely, Transformer-XL~\cite{dai2019transformer} and GPT-2~\cite{radford2019language}. We believe decoder parameter count should be regarded as a competitive baseline for evaluating Transformer NAS, both in terms of ranking capabilities and easy computation. We open-source our code along with tabular information of our trained models to foster future NAS research on Transformers.




%% file: 2_related_work.tex
\vspace{-0.2cm}
\section{Related Work}\label{sec:related_work}
\vspace{-0.2cm}
Here, we discuss literature on automated search for Transformer architectures in the language domain. We refer to extensive surveys on NAS~\citep{elsken2019neural,wistuba2019survey} for a broader overview
of the field. 

\noindent \textbf{Decoder-only Architectures.} \citet{primer} search over TensorFlow programs that implement an autoregressive language model via evolutionary search. 
Since most random sequences of programs either
have errors or underperform, the search has to be seeded with the 
regular Transformer architecture, termed ``Primer''. As opposed to ``Primer'' which 
uses large computation to search a general space, we aim 
to efficiently search the ``backbone'' of traditional 
decoder-only Transformers. Additionally, the objective in ``Primer'' is to find models that train faster. Our objective for NAS, however, is to deliver Pareto-frontiers for inference, with respect to perplexity and hardware constraints. 

\noindent \textbf{Encoder-only Architectures.} Relative to decoder-only 
autoregressive language models, encoder-only architectures
like BERT~\cite{devlin2019bert} have received much more recent attention from the NAS community. NAS-BERT~\cite{nasbert} trains a supernet to efficiently search for
masked language models (MLMs) which are compressed versions of the standard BERT,
Such models can then be used in downstream 
tasks as is standard practice. 
Similar to NAS-BERT,~\citet{analyzingnas} train 
a supernet to conduct architecture 
search with the aim of finding more efficient BERT variants. They find interesting 
empirical insights into supernet training issues like differing gradients at the same 
node from different child architectures and different tensors as input and output 
at every node in the supernet. The authors propose fixes
that significantly improve supernet training. \citet{fasttransformers,autotinybert,autobertzero}
also conduct variants of supernet training with the aim of finding
more efficient BERT models.

\noindent \textbf{Encoder-Decoder Related:} Applying the well-known DARTS~\cite{liu2018darts} approach to 
Transformer search spaces leads to memory-hungry supernets. To mitigate this issue, \citet{memoryefficientsearch} propose a multi-split reversible network and a memory-efficient
backpropagation algorithm. One of the earliest papers 
that applied discrete NAS to Transformer search spaces
was \cite{evolvedtransformer}, which uses a modified form of evolutionary search. 
Due to the expense of directly performing discrete search on the search space, 
this work incurs extremely large computation overhead.
Follow-up work by~\cite{wang2020hat} uses the Once-For-All~\cite{ofa} approach to train a supernet for encoder-decoder architectures used in machine translation. Search is performed on subsamples of the supernet that inherit weights to estimate task accuracy. For each target device, the authors train a small neural network regressor on thousands of architectures to estimate latency.
As opposed to using a latency estimator, \sys{} evaluates the latency of each candidate architecture on the target hardware. Notably, by performing the search directly on the target platform, \sys{} can easily incorporate various hardware performance metrics, e.g., peak memory utilization, for which accurate estimators may not exist. To the best of our knowledge, such holistic integration of multiple hardware metrics in Transformer NAS has not been explored previously.










%% file: 3_methodology.tex
\vspace{-0.3cm}
\section{Lightweight Transformer Search}\label{sec:methodology}
\vspace{-0.2cm}
We perform an evolutionary search over candidate architectures
to extract models that lie on the Pareto-frontier. In contrast to the vast majority of prior methods that deliver a single architecture from the search space, our search is performed over the entire Pareto, generating architectures with a wide range of latency, peak memory utilization, and perplexity with one round of search. This alleviates the need to repeat the NAS algorithm for each hardware performance constraint.

To evaluate candidate models during the search, \sys{} uses a training-free proxy for the validation perplexity. 
By incorporating training-free evaluation metrics, \sys{}, for the first time, performs the entire search directly on the target (constrained) hardware. Therefore, we can use real measurements of hardware performance during the search.
Algorithm~\ref{alg:algorithm} outlines the iterative process
\begin{wrapfigure}{r}{0.55\textwidth}
\vspace{-0.3cm}
\setlength{\algomargin}{8pt}
\begin{algorithm}[H]
\small
\DontPrintSemicolon
\SetNoFillComment
\SetAlgoLined
\KwIn{Search space $\mathcal{D}$, $n_{iter}$}
\KwOut{Perplexity-latency-memory Pareto-frontier $\mathbb{F}$}
$\mathcal{L}, \mathcal{M}, \mathcal{P}, \mathbb{F} \gets \emptyset, \emptyset,\emptyset, \emptyset$\;
\While{$N \leq n_{iter}$}{
    $\mathbb{F}' \gets$ Subsample($\mathbb{F}$)\;
    $\mathbb{S}_N \gets EA(\mathbb{F}', \mathcal{D})$\;
    \tcp{hardware profiling}
    $\mathcal{L} \gets \mathcal{L} \bigcup$ 
    Latency($\mathbb{S}_N$)\;
    $\mathcal{M} \gets \mathcal{M} \bigcup$ 
    Memory($\mathbb{S}_N$)\;
    \tcp{estimate perplexity}
    $\mathcal{P} \gets \mathcal{P} \bigcup$ Proxy($\mathbb{S}_N$)\;
    \tcp{update the Pareto-frontier}
    $\mathbb{F} \gets $ LowerConvexHull($\mathcal{P}, \mathcal{L}, \mathcal{M}$)
}
\caption{\sys{}'s training-free NAS}
\label{alg:algorithm}
\end{algorithm}
\vspace{-0.3cm}
\end{wrapfigure}
performed in \sys{} \footnote{The Pareto-frontier search method in Algorithm \ref{alg:algorithm} is inspired by \citep{elsken2018lemonade} and \citep{petridish2019}. Other possibilities include variations proposed in \citep{guerrero2021bagofbaselines}, evaluation of which is orthogonal to our contributions in this work.} for finding candidate architectures in the search space ($\mathcal{D}$), that lie on the $3$-dimensional Pareto-frontier ($\mathbb{F}$) of perplexity versus latency and memory. At each iteration, a set of points ($\mathbb{F'}$) are subsampled from the current Pareto-frontier. A new batch of architectures ($\mathbb{S}_N$) are then sampled from $\mathbb{F'}$ using evolutionary algorithms ($EA(.)$). The new samples are evaluated in terms of latency ($\mathcal{L}$), peak memory utilization ($\mathcal{M}$), and validation perplexity ($\mathcal{P}$). Latency and memory are measured directly on the target hardware while the perplexity is indirectly estimated using our accurate and training-free proxy methods. Finally, the Pareto-frontier is recalibrated 
using the lower convex hull of all sampled architectures. In the context of multi-objective NAS, Pareto-frontier points are those where no single metric (e.g., perplexity, latency, and memory) can be improved without degrading at least one other metric~\cite{guerrero2021bagofbaselines}.
To satisfy application-specific needs, optional upper bounds can be placed on the latency and/or memory of sampled architectures during search.

\begin{wrapfigure}{r}{0.48\textwidth}\vspace{-0.3cm}
    \centering
    \vspace{-0.2cm}
    \includegraphics[width=0.37\columnwidth]{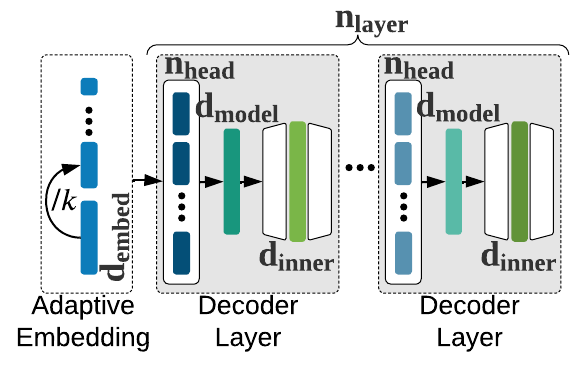}
    \vspace{-0.2cm}
    \caption{Elastic parameters in \sys{} search space.}\label{fig:search_space}
\vspace{-0.2cm}
\end{wrapfigure}

\noindent\textbf{Search Space.}
Figure~\ref{fig:search_space} shows all elastic parameters in \sys{} search space, namely, number of layers (n\textsubscript{layer}), number of attention heads (n\textsubscript{head}), decoder output dimension (d\textsubscript{model}), inner dimension of the feed forward network (d\textsubscript{inner}), embedding dimension (d\textsubscript{embed}), and the division factor ($k$)  of adaptive embedding~\cite{baevski2018adaptive}. These architectural parameters are compatible with popularly used autoregressive
 Transformer backbones, e.g., GPT. For preliminaries on autoregressive Transformers, please see Appendix~\ref{sec:appdx_prelim}. We adopt a  heterogeneous search space where the backbone parameters are decided on a per-layer basis. This is in contrast to the homogeneous structure commonly used in Transformers~\cite{dai2019transformer,brown2020language}, which reuses the same configuration for all layers. Compared to homogeneous models, the flexibility associated with heterogeneous architectures enables them to obtain much better hardware performance under the same perplexity budget (see Section~\ref{sec:comparisons}). 

Heterogeneous search space was previously explored in~\cite{wang2020hat}. However, due to the underlying supernet structure, not all design parameters can change freely. As an example, the dimensionality of the \textit{Q, K, V} vectors inside the encoder and decoder layers is fixed to a large value of $512$ to accommodate inheritance from the supernet. 
Our search space, however, allows exploration 
of all internal dimensions without constraints. By not relying on the supernet structure, our search space easily encapsulates various Transformer backbones with different configurations of the input/output embedding layers and elastic internal dimensions. 

\sys{} searches over the following values for the architectural parameters in our backbones: n\textsubscript{layer}$\in \{2,\dots, 16|1\}$\footnote{We use the notation \{v\textsubscript{min},\dots, v\textsubscript{max}|step size\} to show the valid range of values.}, d\textsubscript{model}$\in \{128,\dots,1024|64\}$, d\textsubscript{inner}$\in \{256,\dots,4096|64\}$, and n\textsubscript{head}$\in \{2,4,8\}$. Additionally we explore adaptive input embedding~\cite{baevski2018adaptive} with d\textsubscript{embed}$\in \{128, 256, 512\}$ and factor $k\in\{1,2,4\}$. 
Once a d\textsubscript{model} is sampled, we adjust the lower bound of the above range for d\textsubscript{inner} to $2\times$d\textsubscript{model}. Encoding this heuristic inside the search ensures that the acquired models will not suffer from training collapse.
Our heterogeneous search space encapsulates more than $10^{54}$ different architectures. Such high dimensionality further validates the critical need for training-free NAS. 


\vspace{-0.2cm}
\subsection{Training-free Architecture Ranking}\label{sec:proxies}

\input{3_2_1_pruning}
\input{3_2_2_decoder_params}

\comment{
\vspace{-0.3cm}
\subsection{Evolutionary Search}
We use evolutionary algorithms to search over the candidate Transformer architectures. In contrast to the vast majority of prior methods that deliver a single architecture from the search space, our search is performed over the entire Pareto. As a result, \sys{} generates architectures with a wide range of latency, peak memory utilization, and perplexity characteristics with one round of search. This alleviates the need to repeat the NAS algorithm for each hardware performance constraint. At each iteration of the evolutionary search, we utilize our training-free proxy to rank the architectures in terms of their validation perplexity and measure the latency and peak memory utilization directly on the target hardware. We then estimate the Pareto-frontier of perplexity versus latency and memory using the lower convex hull of all sampled architectures. In the context of multi-objective NAS, Pareto-frontier points are those where no single metric (e.g., perplexity, latency, and memory) can be improved without degrading at least one other metric~\cite{guerrero2021bagofbaselines}.
The models on the Pareto-frontier are then used to sample new architectures using evolutionary operations, i.e., mutation and crossover. To satisfy application-specific needs, optional upper bounds can be placed on the latency and/or memory of sampled architectures during the search.
}

%% file: 3_2_1_pruning.tex
\noindent\textbf{$\blacktriangleright$ Low-cost Ranking Proxies.}
Recently, \citet{abdelfattah2020zero} utilize the summation of pruning scores over all model
weights as the ranking proxy for Convolutional Neural Networks (CNNs), where a higher score corresponds to higher architecture rank in the search space. \citet{colin2022adeeperlook} analyze these and more recent proxies and find that no particular proxy performs consistently well over various tasks and baselines, while parameter and floating point operations (FLOPS) count proxies are quite competitive. However, they did not include Transformer-based search spaces in their analysis. 
To the best of our knowledge, low-cost (pruning-based) proxies have not been evaluated on Transformer search spaces in the  language domain. Note that one cannot naively apply these proxies to language models. Specifically, since the embedding layer in Transformers is equivalent to a lookup operation, special care must be taken to omit this layer from the proxy computation. Using this insight, we perform the first systematic study of low-cost proxies for NAS on autoregressive Transformers for text prediction. 

We leverage various pruning metrics, namely, \gradnorm, \snip~\cite{lee2018snip}, \grasp~\cite{wang2020picking}, \fisher~\cite{theis2018faster}, and \synflow~\cite{tanaka2020pruning}. We also study \jacobcov~\cite{mellor2020neural} and
\relulogdet~\cite{mellor2021neural} which are low-cost scoring mechanisms proposed for NAS on CNNs in vision tasks. 
While these low-cost techniques do not perform model training, they require forward and backward passes over the architecture to compute the proxy, which can be time-consuming on low-end hardware. Additionally, the aforesaid pruning techniques, by definition, incorporate the final softmax projection layer in their score assessment. Such an approach seems reasonable for CNNs dealing with a few classification labels, however, it can skew the evaluation for autoregressive Transformers dealing with a large output vocabulary space. 
To overcome these shortcomings, we introduce a zero-cost architecture ranking strategy in the next section that outperforms the proposed low-cost proxies in terms of ranking precision, is data free, and does not perform any forward/backward propagation.



%% file: 3_2_2_decoder_params.tex
\begin{figure}[t]
    \centering
    \includegraphics[width=0.8\columnwidth]{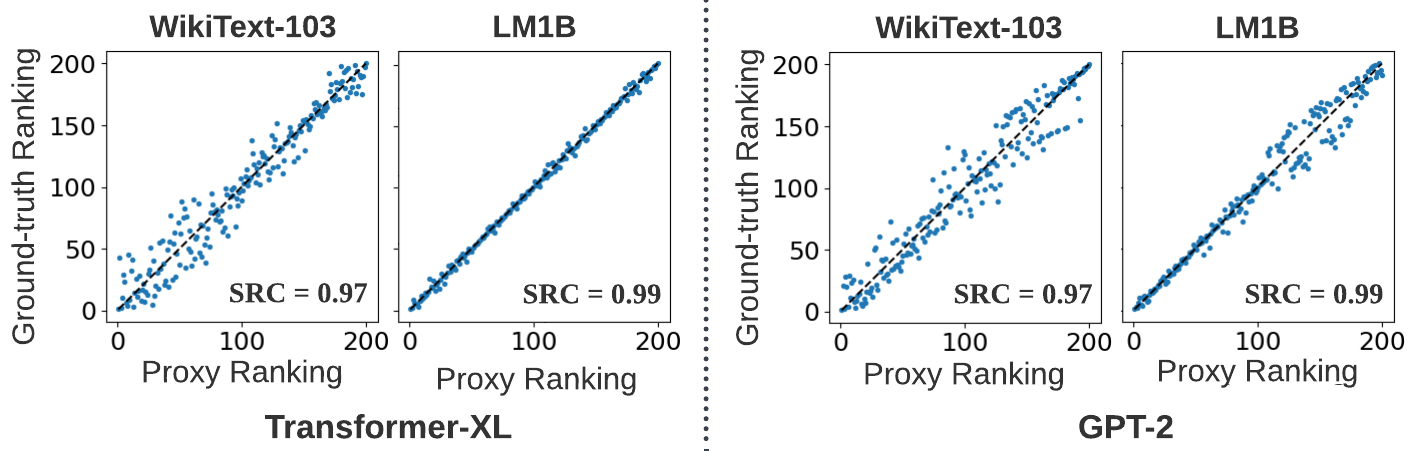}
    \vspace{-0.2cm}
    \caption{Our training-free zero-cost proxy based on decoder parameter count is highly correlated with the (ground truth) validation perplexity after full training. Each plot contains $200$ architectures sampled randomly from the search space of Transformer-XL or GPT-2 backbone.}\label{fig:zero_proxy}
\vspace{-0.6cm}
\end{figure}

\noindent\textbf{$\blacktriangleright$ Decoder Parameter Count as a Proxy.} We empirically establish a strong correlation between the parameter count of decoder layers and final model performance in terms of validation perplexity. 
We evaluate $200$ architectures sampled uniformly at random from the search space of two autoregressive Transformer backbones, namely, Transformer-XL and GPT-2. These architectures are trained fully on WikiText-103 and One Billion Word (LM1B) datasets, which
consumes over $\mathbf{25000}$ GPU-hours on NVIDIA A100 and V100 nodes. 
We compare the ranking obtained using decoder parameter count proxy and the ground truth ranking after full training in Figure~\ref{fig:zero_proxy}. On WikiText-103, zero-cost ranking using the decoder parameter count obtains a Spearman's Rank Correlation (SRC) of $0.97$ with full training. SRC further increases to $0.99$ for the more complex LM1B benchmark on both backbones. This validates that the decoder parameter count is strongly correlated with final model performance, thereby providing a reliable training-free proxy for NAS.

%% file: 4_experiments.tex
\vspace{-0.3cm}
\section{Experiments}\label{sec:experiments}
\vspace{-0.1cm}
We conduct experiments to seek answers to the following critical questions:

\noindent\circled{1} How well can training-free proxies perform compared to training-based methods for estimating the performance of Transformer models?

\noindent\circled{2} How does model topology affect the performance of the proposed decoder parameter proxy?

\noindent\circled{3} Can our training-free decoder parameter count proxy be integrated inside a search algorithm to estimate the Pareto-frontier? How accurate is such an estimation of the Pareto?

\noindent\circled{4} Which models are on the Pareto-frontier of perplexity, latency, and memory for different hardware?

\noindent\circled{5} How well do \sys{} models perform in zero and one-shot settings compared to hand-designed variants when evaluated on downstream tasks?

We empirically answer questions \circled{1}, \circled{2}, \circled{4}, and \circled{5} in Sections~\ref{sec:exp_proxies}, \ref{sec:topology}, \ref{sec:comparisons}, and~\ref{sec:opt} respectively. We further address question \circled{3} in Appendix~\ref{sec:nas} where we show the Pareto-frontier models extracted by the decoder parameter count proxy are very close to the ground truth Pareto-frontier with an average of $0.6\%$ perplexity difference. Additionally, we show the efficacy of the decoder parameter count proxy when performing search on different ranges of model sizes in Appendix~\ref{sec:nas}, Figure~\ref{fig:param_count_bins}.

\vspace{-0.2cm}
\subsection{Experimental Setup}\label{sec:setup}
Please refer to Appendix~\ref{sec:appdx_training} for information about the benchmarked datasets, along with details of our training and evaluation setup, hyperparameter optimization, and evolutionary search algorithm.

\noindent\textbf{Backbones.} We apply our search on two widely used autoregressive Transformer backbones, namely, Transformer-XL~\cite{dai2019transformer} and GPT-2~\cite{radford2019language} that are trained from scratch with varying architectural hyperparameters. The internal structure of these backbones are quite similar, containing decoder blocks with attention and feed-forward layers. The difference between the backbones lies mainly in their dataflow structure; the Transformer-XL backbone adopts a recurrence methodology over past states coupled with relative positional encoding which enables modeling longer term dependencies. 

\comment{
\noindent\textbf{Datasets.} 
We conduct experiments on two datasets, WikiText-103~\cite{merity2016pointer} and LM1B~\cite{chelba2013one}. The datasets are tokenized
using word-level and byte-pair encoding for models with Transformer-XL and GPT-2 backbones, respectively.

\noindent\textbf{Training and Evaluation.}
We adopt the open-source code by~\cite{transformer_github} and~\cite{huggingface} to implement the Transformer-XL and GPT-2 backbones, respectively. 
For each backbone and dataset, we use the same training setup for all models generated by NAS. We provide details of our training hyperparameters in Table~\ref{tab:training_hyperparams} of the supplementary material.
Validation perplexity is measured over a sequence length of $192$ and $32$ tokens for WikiText-103 and LM1B datasets, respectively. Inference latency and peak memory utilization are measured on the target hardware for the sequence length, averaged over $10$ measurements. We utilize PyTorch's native benchmarking interface for measuring the latency and memory utilization of candidate architectures.
}

\noindent \textbf{Performance Criteria.} To evaluate the ranking performance of various proxies, we first establish a ground truth ranking of candidate architectures by training them until full convergence. This ground truth ranking is then utilized to compute two performance criteria as follows:

\noindent$\blacktriangleright$ \underline{Common Ratio (CR):} We define CR as the percentage overlap between the ground truth ranking of architectures versus the ranking obtained from the proxy. CR quantifies the ability of the proxy ranking to identify the top$k\%$ architectures based on their validation perplexity after full training.

\noindent$\blacktriangleright$ \underline{Spearman's Rank Correlation (SRC):} We use this metric to measure the correlation between the proxy ranking and the ground truth. Ideally, the proxy ranking should have high correlation with the ground truth over the entire search space as well as high-performing candidate models.




\comment{
\noindent \textbf{Search Setup.}
Evolutionary search is performed for $30$ iterations with a population size of $100$; the parent population accounts for $20$ samples out of the total $100$; $40$ mutated samples are generated per iteration from a mutation probability of $0.3$; and $40$ samples are created using crossover.
}

\input{4_1_ranking_proxy}
\input{4_3_Comparisons}
\input{4_4_opt}

%% file: 4_1_ranking_proxy.tex
\vspace{-0.2cm}
\subsection{How do training-free proxies perform compared to training-based methods?}\label{sec:exp_proxies}
\vspace{-0.1cm}
In this section, we benchmark several proxy methods for estimating the rank of candidate architectures. Specifically, we investigate three different ranking techniques, namely, partial training, low-cost methods, and number of decoder parameters.

\begin{figure}[t]
    \centering
    \includegraphics[width=0.99\textwidth]{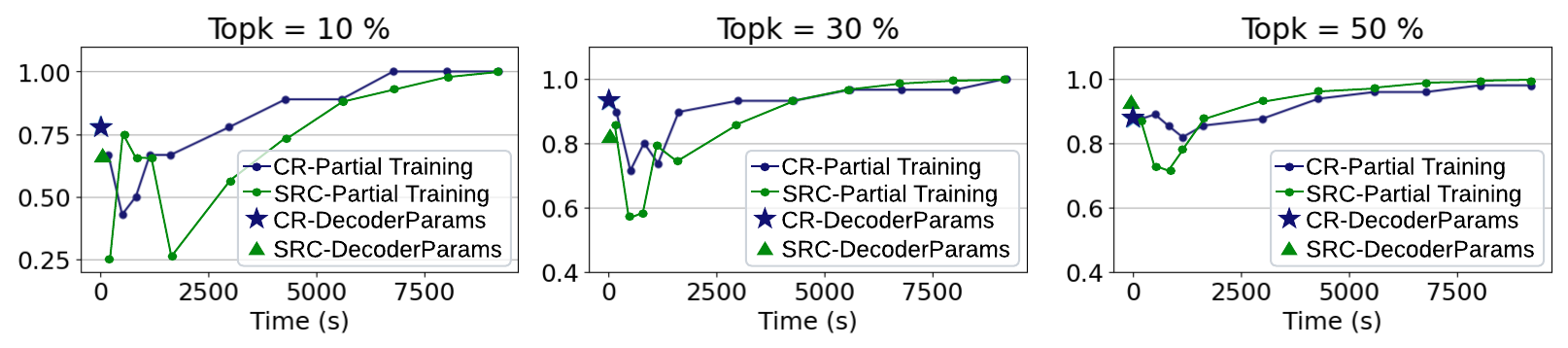}
    \vspace{-0.2cm}
    \caption{Comparison between partial training and our zero-cost proxy, i.e., decoder parameter count, in terms of ranking performance and timing overhead. Each subplot corresponds to a top$k\%$ of the randomly sampled models, based on their validation perplexity after full training.
}\label{fig:proxies}
\vspace{-0.5cm}
\end{figure}

\noindent $\blacktriangleright$ \textbf{Partial Training.} We first analyze the relationship between validation perplexity after a shortened training period versus that of full training for ranking candidate models. We stop the training after $\tau \in[1.25\%, 87.5\%]$ of the total training iterations needed for model convergence. Figure~\ref{fig:proxies} demonstrates the SRC and CR of partial training with various $\tau$s, evaluated on $100$ randomly selected models from the Transformer-XL backbone, trained on WikiText-103. The horizontal axis denotes the average time required for $\tau$ iterations of training across all sampled models.
Intuitively, a higher number of training iterations results in a more accurate estimate of the final perplexity. Nevertheless, the increased wall-clock time prohibits training during search and also imposes the need for GPUs. Interestingly, very few training iterations, i.e., $1.25\%$, provide a very good proxy for final performance with an SRC of $>0.9$ on the entire population. Our training-free proxy, i.e., decoder parameter count, also shows competitive SRC compared to partial training.


\begin{wrapfigure}{r}{0.5\textwidth}
    \centering
    \vspace{-0.3cm}
    \includegraphics[width=0.49\columnwidth]{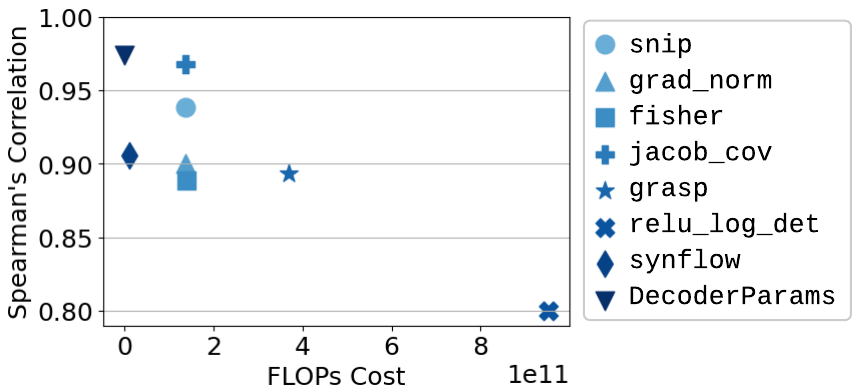}
    \vspace{-0.2cm}
    \caption{SRC between low-cost proxies and the ground truth ranking after full training of 200 randomly sampled Transformers. The decoder parameter count obtains the best SRC with zero cost.}\vspace{-0.3cm}
    \label{fig:pruning}
\end{wrapfigure}

\noindent $\blacktriangleright$ \textbf{Low-cost Proxies.} We benchmark various low-cost methods introduced in Section~\ref{sec:proxies} on $200$ randomly sampled architectures from the Transformer-XL
backbone, trained on WikiText-103. 
Figure~\ref{fig:pruning} shows the SRC between low-cost proxies and the ground truth  ranking after full training.
We measure the cost of each proxy in terms of FLOPs. 
As seen, the evaluated low-cost proxies have a strong correlation with the ground truth ranking (even the lowest performing \relulogdet has $>0.8$ SRC), validating the effectiveness of training-free NAS on autoregressive Transformers. 
The lower performance of \relulogdet can be attributed to the much higher frequency of ReLU activations in CNNs, for which the method was originally developed, compared to Transformer-based architectures.  
Our analysis of randomly selected models with homogeneous structures also shows a strong correlation between the low-cost proxies and validation perplexity, with
decoder parameter count outperforming other proxies (see Appendix~\ref{sec:appdx_homogeneous}).

\begin{figure}[t]
    \centering
    \includegraphics[width=0.99\columnwidth]{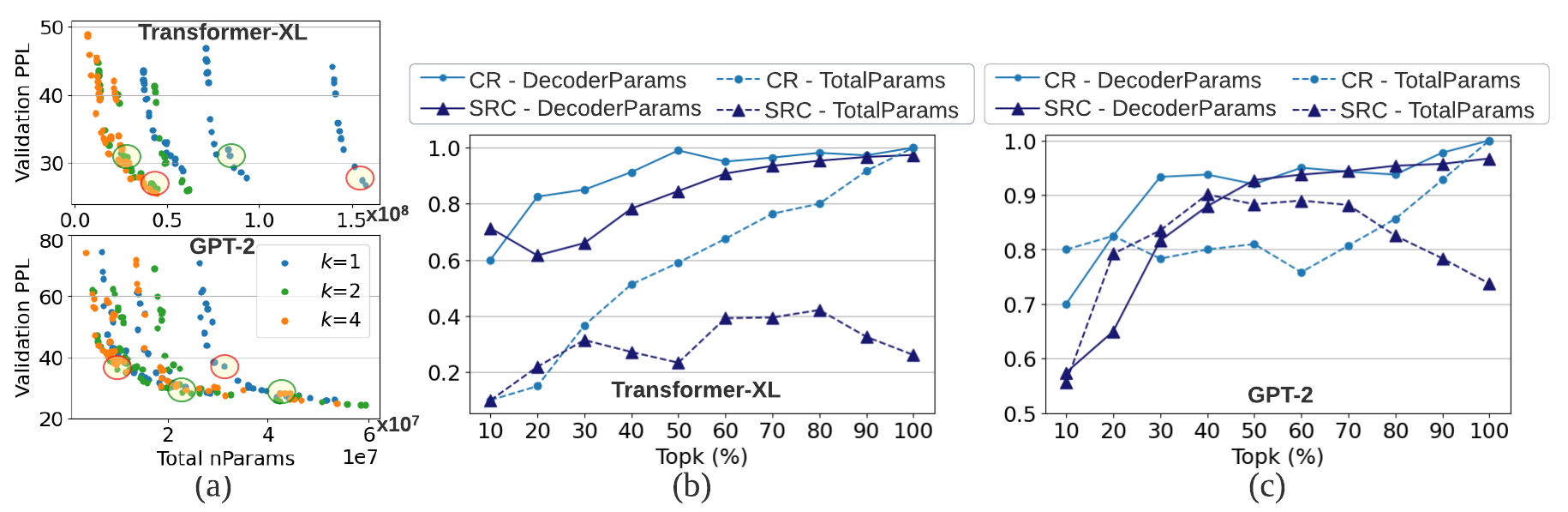}
     \caption{(a) Validation perplexity after full training versus total parameters for $200$ randomly sampled architectures trained on WikiText-103. The clear downward trend suggests a strong correlation between parameter count and perplexity. (b), (c) Performance of parameter count proxies for ranking the randomly sampled architectures with Transformer-XL and GPT-2 backbones.
    }\label{fig:param_proxy}
\vspace{-0.3cm}
\end{figure}

\noindent $\blacktriangleright$ \textbf{Parameter Count.} Figure~\ref{fig:param_proxy}a demonstrates the final validation perplexity versus the total number of model parameters for $200$ randomly sampled architectures from GPT-2 and Transformer-XL backbones.
This figure contains two important observations: (1)~the validation perplexity has a downward trend as the number of parameters increases, (2)~The discontinuity is caused by the dominance of embedding parameters when moving to the small Transformer regime. We highlight several example points in Figure~\ref{fig:param_proxy}a where the architectures are nearly identical but the adaptive input embedding factor $k$ is changed. Changing $k\in\{1,2,4\}$ (shown with different colors in Figure~\ref{fig:param_proxy}a) varies the total parameter count without much influence on the validation perplexity. 

The above observations 
motivate us to evaluate two proxies, i.e., total number of parameters and decoder parameter count. 
Figures~\ref{fig:param_proxy}b and~\ref{fig:param_proxy}c demonstrate the CR and SRC metrics evaluated on the $200$ randomly sampled models divided into top$k\%$ bins based on their validation perplexity. As shown, the total number of parameters generally has a lower SRC with the validation perplexity, compared to decoder parameter count. This is due  to the masking effect of embedding parameters, particularly in the Transformer-XL backbone. The total number of decoder parameters, however, provides a highly accurate, zero-cost proxy with an SRC of $0.97$ with the perplexity over all models, after full training. We further show the high correlation between decoder parameter count and validation perplexity for Transformer architectures with homogeneous decoder blocks in the supplementary material, Appendix~\ref{sec:appdx_homogeneous}. While our main focus is on autoregressive, decoder-only, Transformers, we provide preliminary results on the ranking performance of parameter count proxies for encoder-only and encoder-decoder Transformers in Appendix~\ref{sec:appdx_domains}.

\vspace{-0.2cm}
\subsection{How does variation in model topology affect decoder parameter count as a proxy?}\label{sec:topology}
\vspace{-0.2cm}

The low-cost proxies introduced in Section~\ref{sec:proxies}, rely on forward and backward passes through the network. As such, they automatically capture the topology of the underlying architecture via the dataflow. The decoder parameter count proxy, however, is topology-agnostic. In this section, we investigate the effect of topology on the performance of decoder parameter count proxy. Specifically, we seek to answer whether for a given decoder parameter count budget, the aspect ratio of the architecture, i.e., trading off the width versus the depth, can affect the final validation perplexity.  

\begin{figure*}[t]
\vspace{-0.2cm}
    \centering
    \begin{subfigure}[b]{0.32\textwidth}
     \centering
    \includegraphics[width=\columnwidth]{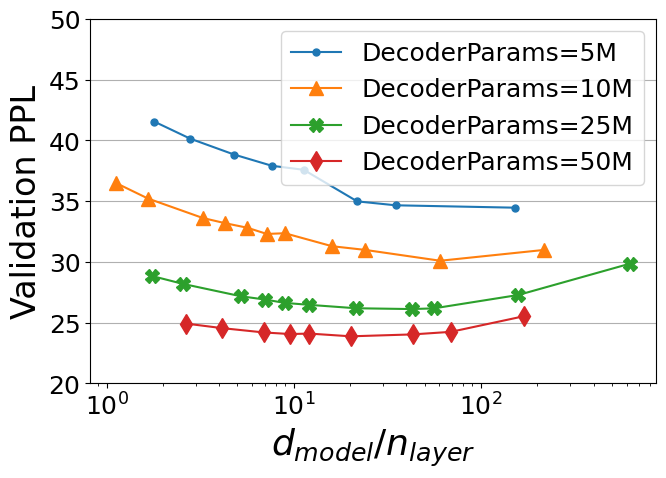}
    \caption{}\label{fig:scaling_exps_gpt_a}
    \end{subfigure}
    \begin{subfigure}[b]{0.32\textwidth}
     \centering
     \includegraphics[width=\columnwidth]{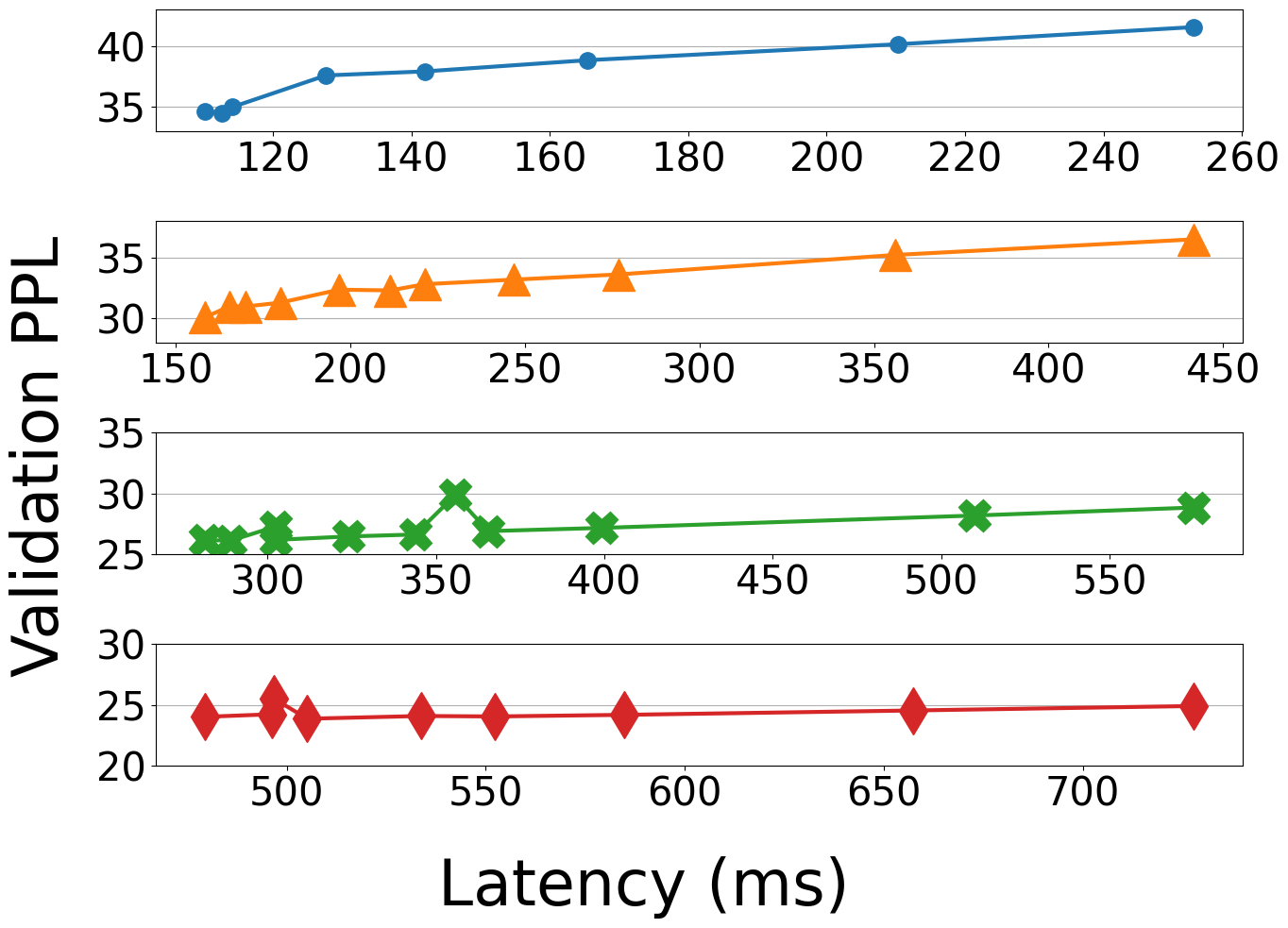}
    \caption{}\label{fig:scaling_exps_gpt_b}
    \end{subfigure}
    \begin{subfigure}[b]{0.32\textwidth}
     \centering
     \includegraphics[width=\columnwidth]{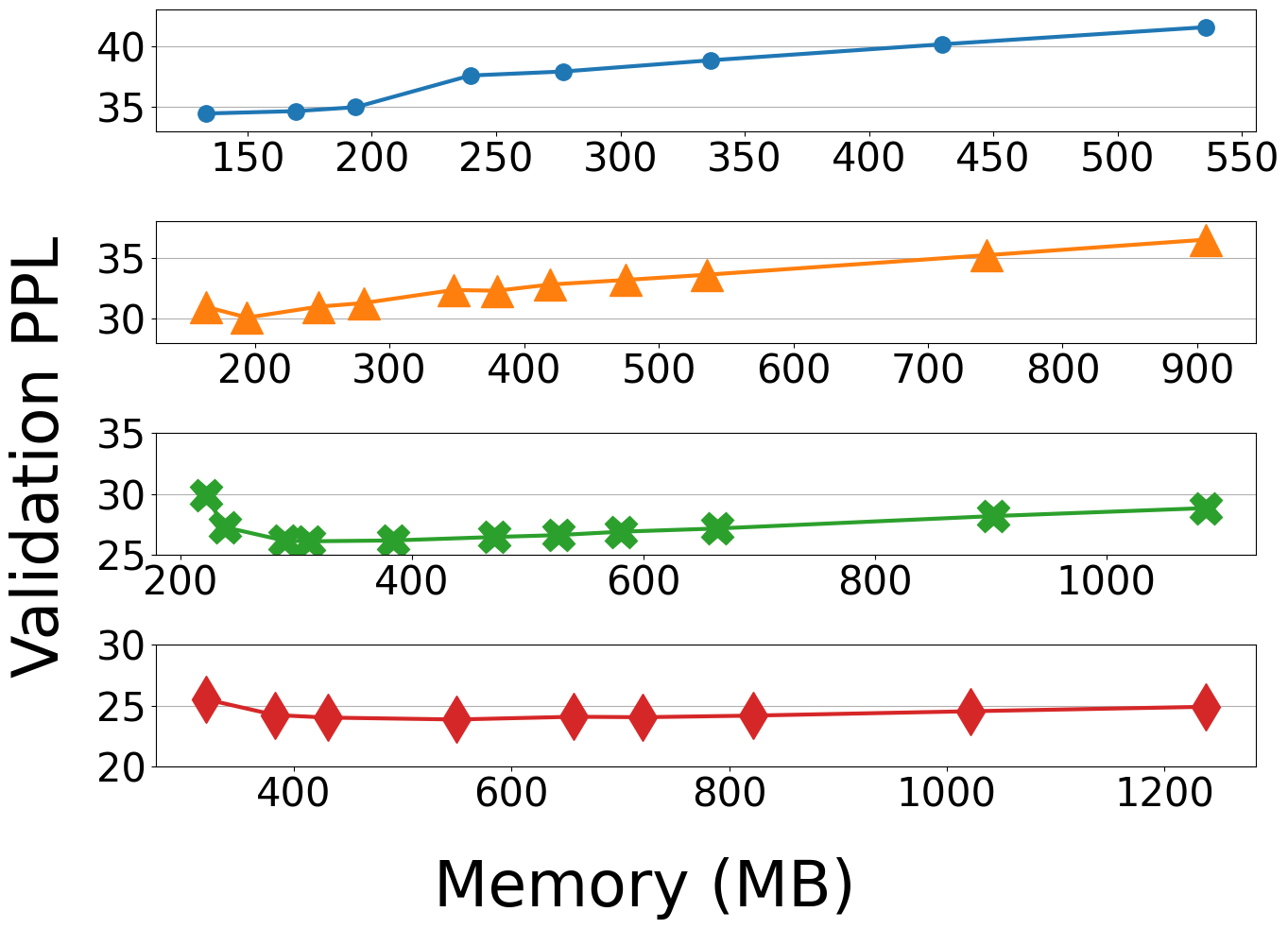}
    \caption{}\label{fig:scaling_exps_gpt_c}
    \end{subfigure}
    \vspace{-0.2cm}
    \caption{Validation perplexity after full training versus the (a) width-to-depth aspect ratio, (b) latency, and (c) peak memory utilization. Models are randomly generated from the GPT-2 backbone and trained on WikiText-103. For a given decoder parameter count, we observe low variation in perplexity across different models, regardless of their topology. The topology, however, significantly affects the latency (up to $2.8\times$) and peak memory utilization (up to $5.5\times$) for models with the same perplexity.}
    \label{fig:scaling_exps_gpt}
    \vspace{-0.5cm}
\end{figure*}

We define the aspect ratio of the architecture as d\textsubscript{model} (=width), divided by n\textsubscript{layer} (=depth). This metric provides a sense of how skewed the topology is and has been used in prior works which study scaling laws for language models \citep{kaplan2020scaling}. For a given decoder parameter count budget, we generate several random architectures from the GPT-2 backbone with a wide range of the width-to-depth aspect ratios\footnote{We control the aspect ratio by changing the width, i.e., d\textsubscript{model} while keeping d\textsubscript{inner}=$2\times$d\textsubscript{model} and n\textsubscript{head}=$8$. The number of layers is then derived such that the total parameter count remains the same.}. The generated models span wide, shallow topologies (e.g., d\textsubscript{model}=$1256$, n\textsubscript{layer}=$2$) to narrow, deep topologies (e.g., d\textsubscript{model}=$112$, n\textsubscript{layer}=$100$). Figure~\ref{fig:scaling_exps_gpt_a} shows the validation perplexity of said architectures after full training on WikiText-103 versus their aspect ratio. The maximum deviation (from the median) of the validation perplexity is $<12.8\%$ for a given decoder parameter count, across a wide range of aspect ratios $\in[1,630]$. Our findings on the heterogeneous search space complement the empirical results by \cite{kaplan2020scaling} where decoder parameter count largely determines perplexity for homogeneous Transformer architectures, irrespective of shape (see Figure 5 in \citep{kaplan2020scaling}).

We observe stable training when scaling models from the GPT-2 backbone up to $100$ layers, with the perplexity increasing only when the aspect ratio nears $1$. Nevertheless, such deep models are not part of our search space as they have high latency and are unsuitable for lightweight inference. For the purposes of hardware-aware and efficient Transformer NAS, decoder parameter count proxy holds a very high correlation with validation perplexity, regardless of the architecture topology as shown in Figure~\ref{fig:scaling_exps_gpt_a}. 
We further validate the effect of topology on decoder parameter count proxy for the Transformer-XL backbone in Figure~\ref{fig:scaling_exps} of Appendix~\ref{sec:appdx_scaling}. Our results demonstrate less than $7\%$ deviation (from the median) in validation perplexity for different aspect ratios $\in[8,323]$. 

Note that while models with the same parameter count have very similar validation perplexities, the topology in fact affects their hardware performance, i.e., latency (up to $2.8\times$) and peak memory utilization (up to $5.5\times$), as shown in Figures~\ref{fig:scaling_exps_gpt_b} and~\ref{fig:scaling_exps_gpt_c}. This motivates the need for incorporating hardware metrics in NAS to find the best topology. 


%% file: 4_3_Comparisons.tex
\vspace{-0.2cm}
\subsection{Pareto-frontier models for various hardware platforms}\label{sec:comparisons}


We run LTS on different target hardware and obtain a range of Pareto-optimal architectures with various latency/memory/perplexity characteristics. During search, we fix the adaptive input embedding factor to $k=4$ to search models that are lightweight while ensuring nearly on-par validation perplexity with non-adaptive input embedding.
As the baseline Pareto, we benchmark the Transformer-XL (base) and GPT-2 (small) models with homogeneous layers $\in [1, 16]$. This is because 
the straightforward way to produce architectures of different latency/memory is varying the number of layers (layer-scaling)~\cite{vaswani2017attention,dai2019transformer}. We compare our NAS-generated architectures with layer-scaled backbones and achieve better validation perplexity and/or lower latency and peak memory utilization. All baseline\footnote{The best reported result in the literature for GPT-2 or Transformer-XL might be different based on the specific training hyperparameters, which is orthogonal to our investigation.}
and NAS-generated models are trained using the same setup enclosed in Table~\ref{tab:training_hyperparams} of Appendix~\ref{sec:appdx_training}. 

Figure~\ref{fig:nas_results} shows the Pareto-frontier architectures found by \sys{} versus the layer-scaled baseline. Here, all models are trained on the LM1B dataset (See Figure~\ref{fig:nas_results_wt103} in Appendix~\ref{sec:comp_wt103} for results on WikiText-103). Note that the Pareto-frontier search is performed in a $3$-dimensional space, an example of which is enclosed in Appendix~\ref{sec:appdx_3dpareto}, Figure~\ref{fig:3d_pareto}. For better visualization, in Figure~\ref{fig:nas_results} we plot $2$-dimensional slices of the Pareto-frontier with validation perplexity on the y-axis and one hardware performance metric (either latency or memory) on the x-axis.

As seen, in the low-latency regime, \sys{} consistently finds models that have significantly lower perplexity compared to naive scaling of the baseline Transformer-XL or GPT-2. On the Transformer-XL backbone, \sys{} finds architectures with an average of $19.8\%$ and $28.8\%$ lower latency and memory, while achieving similar perplexity compared to the baseline on ARM CPU. Specifically, the perplexity of the $16$-layer Transformer-XL base can be replicated on the ARM device with a lightweight model that is $1.6\times$ faster and utilizes $1.9\times$ less memory during execution. On the Corei7 CPU, the Pareto-frontier models found by \sys{} are on average $25.8\%$ faster and  consume $30.0\%$ less memory under the same validation perplexity constraint. In this setting, \sys{} finds a model that replicates the perplexity of the $16$-layer Transformer-XL base while achieving $1.7\times$ faster runtime and $1.9\times$ less peak memory utilization. The savings are even higher on the GPU device, where the NAS-generated models achieve the same perplexity as the baseline with average $30.5\%$ lower latency and $27.0\%$ less memory. Specifically, an \sys{} model with the same perplexity as the $16$-layer Transformer-XL base has  $2.5\times$ lower latency and consumes $2.0\times$ less peak memory on TITAN Xp.

On the GPT-2 backbone, NAS-generated models consume on average $11.8\%$ less memory while achieving the same validation perplexity and latency on an ARM CPU. The benefits are larger on Corei7 and TitanXP where the latency savings are $13.8\%$ and $11.9\%$, respectively. The peak memory utilization also decreases by $9.7\%$ and $12.9\%$, on average, compared to the baseline GPT-2s on Corei7 and TITAN Xp. Notably, NAS finds new architectures with the same perplexity as the $16$-layer GPT-2 with $1.3\times$, $1.5\times$ faster runtime and $1.2\times$ lower memory utilization on Corei7 and TITAN Xp.

Our heterogeneous search space allows us to find a better parameter distribution among decoder layers. Therefore, \sys{} delivers architectures with better performance in terms of perplexity, while reducing both latency and memory when compared to the homogeneous baselines.
We provide the architecture of all baseline and \sys{} models shown in Figure~\ref{fig:nas_results} in Tables~\ref{tab:archs_transxl}-\ref{tab:archs_gpt2} of Appendix~\ref{sec:appdx_archs}.

\begin{figure*}[t]
    \centering
    \includegraphics[width=\textwidth]{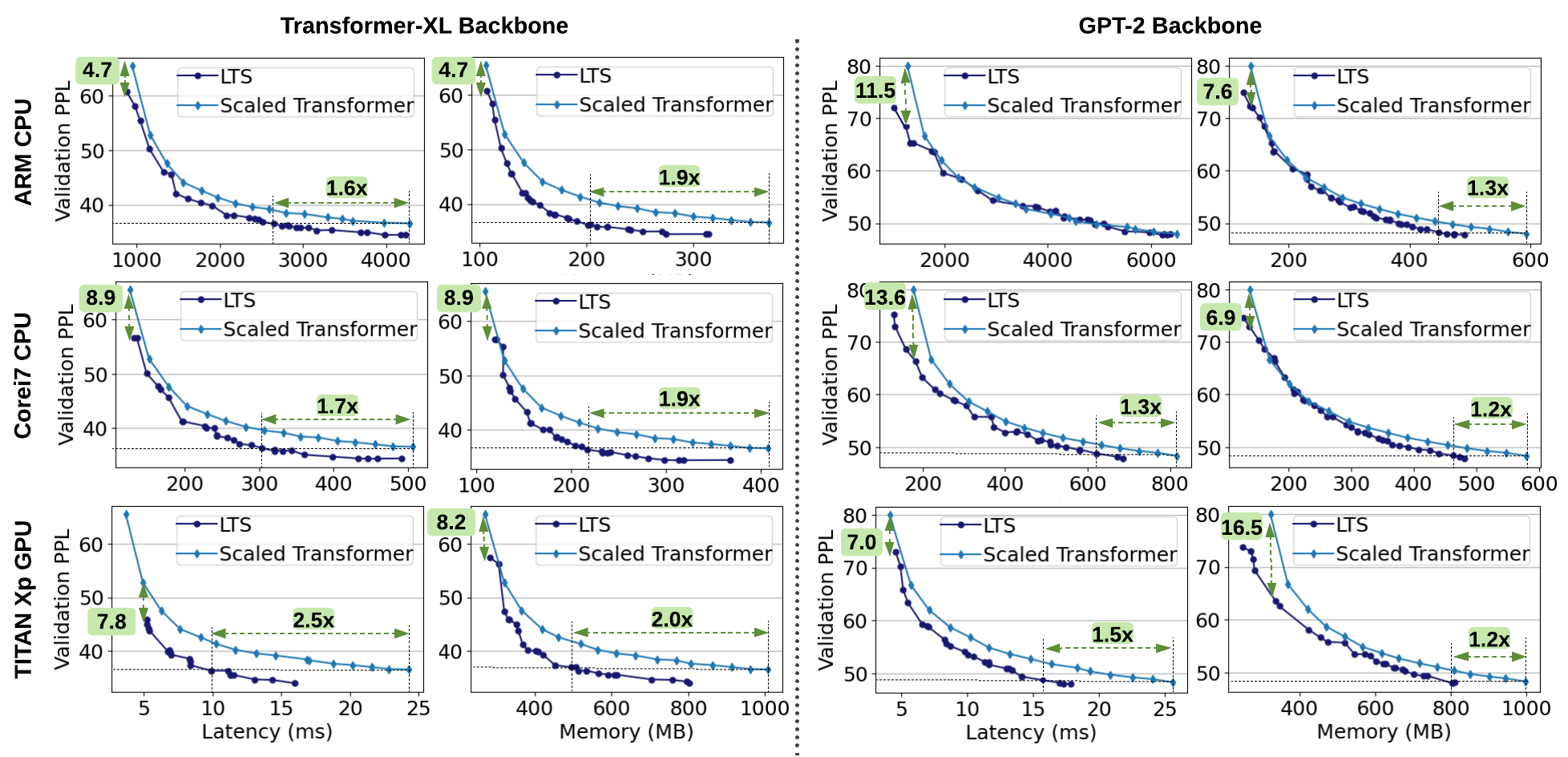}
    \vspace{-0.5cm}
    \caption{2D visualization of the perplexity versus latency and memory Pareto-frontier found by \sys{}, versus the scaled backbone models with varying number of layers, trained on the LM1B dataset. Architectural parameters for models shown here are detailed in Appendix~\ref{sec:appdx_archs}. }
    \label{fig:nas_results}
\vspace{-0.5cm}
\end{figure*}

\noindent $\blacktriangleright$ \textbf{Search Efficiency.}
The main component in \sys{} search time is the latency/peak memory utilization measurement for candidate architectures since evaluating the model perplexity is instant using the decoder parameter count.
Therefore, our search finishes in a few hours on commodity hardware, e.g., taking only $0.9$, $2.6$, and $17.2$ hours on a TITAN Xp GPU, Corei7 CPU, and an ARM core, respectively. To provide more context into the timing analysis, full training of even one $16$-layer Transformer-XL base model on LM1B using a machine with $8\times$ NVIDIA V100 GPUs takes $15.8$ hours. 
Once the Pareto-frontier models are identified, the user can pick a model based on their desired hardware constraints and fully train it on the target dataset. \sys{} is an alternate paradigm to that of training large supernets; our search can run directly on the target device and GPUs are only needed for training the final chosen Pareto-frontier model after search.


In Table~\ref{tab:nas_cost} we study the ranking performance of partial training ($500$ steps) versus the decoder parameter count proxy for evaluating $1200$ architectures from the Transformer-XL backbone during \sys{} search. Astonishingly the decoder parameter count proxy gets higher SRC compared to partial training, while effectively removing training from the inner loop of search for NAS. 

\begin{SCtable}[50][h]
\vspace{-0.3cm}
\centering
\resizebox{0.5\columnwidth}{!}{
\begin{tabular}{lcccc}
\hline
& \begin{tabular}[c]{@{}c@{}}Train\\ Iter\end{tabular}
& \begin{tabular}[c]{@{}c@{}}GPU\\ Hours\end{tabular}
& \begin{tabular}[c]{@{}c@{}}$CO_2$e\\ (lbs)\end{tabular}
& SRC  
\\ \hline \hline
Full Training                     
& 40,000        
& 19,024    
& 5433                                                
& 1.0             
\\ \hline
\multirow{2}{*}{Partial Training} 
& 500           
& 231       
& 66                                                   
& 0.92            
\\ \cline{2-5}
& 5,000         
& 2690     
& 768                                                  
& 0.96            
\\ \hline
\# Decoder Params                 
& \textbf{0}             
& \textbf{0}         
& $\sim$\textbf{0}                                              
& \textbf{0.98}           
\\ \hline
\end{tabular}}
\caption{Ranking abilities of full and partial training versus our proxy for $1200$ models sampled during \sys{} search. Training time is reported for WikiText-103 and NVIDIA V100 GPU. Decoder parameter count proxy obtains an SRC of $0.98$ using zero compute.}\label{tab:nas_cost}
\vspace{-0.5cm}
\end{SCtable}


%% file: 4_4_opt.tex
\begin{SCfigure}[50][t]
    \centering
    \includegraphics[width=0.55\columnwidth]{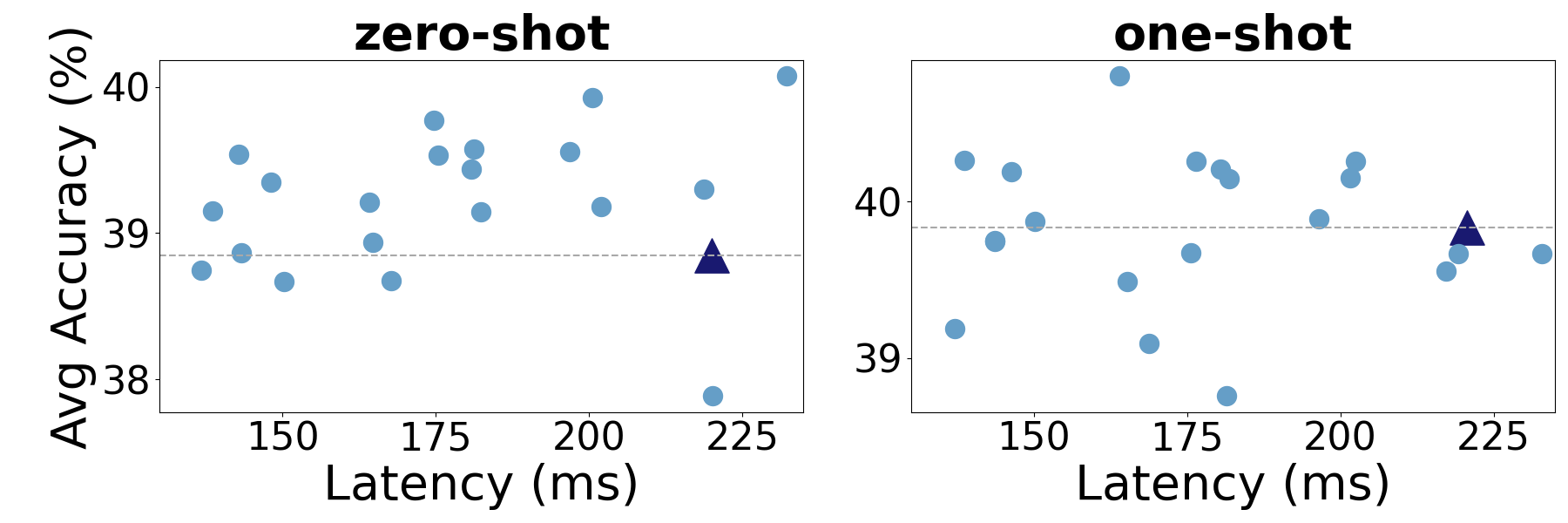}
    \caption{Average zero and one-shot accuracy obtained by \sys{} models (dots) and the baseline OPT-$350$M (triangle) across $14$ NLP tasks. Latency is measured on an A6000 NVIDIA GPU.
    Architectural parameters for all models shown here are detailed in Appendix~\ref{sec:appdx_archs}.}
    \label{fig:avg_acc_opt}
\vspace{-0.5cm}
\end{SCfigure}

\vspace{-0.2cm}
\subsection{Zero and one-shot performance comparison of \sys{} models with OPT}\label{sec:opt}

 \citet{zhang2022opt} open-source a set of pre-trained decoder-only language models, called OPT, which can be used for zero or few-shot inference on various NLP tasks. Below, we compare the performance of \sys{} Pareto-frontier models with the hand-designed OPT architecture in zero and one-shot settings. 
 
We use \sys{} to search for language models with a GPT-2 backbone which have $300$M to $500$M total parameters to compare with the $350$M parameter OPT. Our search space is detailed in Appendix~\ref{sec:appdx_opt}. The search is conducted with latency as the target hardware metric and decoder parameter count as a proxy for perplexity. Once the search concludes, We train $20$ models from the Pareto-frontier along with OPT-$350$M on $28$B tokens from the Pile~\cite{gao2020pile}. The pretrained models are then evaluated on $14$ downstream NLP tasks, namely, HellaSwag~\cite{zellers2019hellaswag}, PIQA~\cite{bisk2020piqa}, ARC (easy and challenge)~\cite{clark2018think}, OpenBookQA~\cite{mihaylov2018can}, WinoGrande~\cite{sakaguchi2021winogrande}, and SuperGLUE~\cite{wang2019superglue} benchmarks BoolQ, CB, COPA, WIC, WSC, MultiRC, RTE, and ReCoRD. The training hyperparameters and the evaluation setup are outlined in Appendix~\ref{sec:appdx_training}.
 Figure~\ref{fig:avg_acc_opt} shows the overall average accuracy obtained across all $14$ tasks versus the inference latency for \sys{} models and the baseline OPT. As shown, NAS-generated models achieve a higher average accuracy with lower latency compared to the hand-designed OPT-$350$M model. We provide a per-task breakdown of zero and one-shot accuracy in Appendix~\ref{sec:appdx_opt}, Figure~\ref{fig:appdx_opt}.
 
\noindent $\blacktriangleright$ \textbf{Zero-shot Performance.} Figure~\ref{fig:appdx_opt_0shot} demonstrates the zero-shot accuracy obtained by \sys{} and OPT-$350$M on the benchmarked tasks. 
Compared to the OPT-$350$M architecture, \sys{} finds models that achieve higher accuracy and lower latency in the zero-shot setting on all evaluated downstream tasks. 
Specifically, the maximum achievable accuracy of our NAS-generated models is $0.2-8.6\%$ higher than OPT-$350$M with an average speedup of $1.2\times$. If latency is prioritized, \sys{} delivers models which are, on average, $1.5\times$ faster and up to $4.6\%$ more accurate than OPT-$350$M . 

\noindent $\blacktriangleright$ \textbf{One-shot Performance.}
Similar trends can be observed for one-shot evaluation as shown for different tasks in Figure~\ref{fig:appdx_opt_1shot}. \sys{} Pareto-frontier models improve the per-task accuracy of OPT-$350$M on $12$ out of $14$ tasks by $0.1-8.0\%$, while achieving an average speedup of $1.2\times$. On the same tasks, \sys{} Pareto-frontier includes models that enjoy up to $1.6\times$ speedup over OPT-$350$M with an average $1.5\%$ higher accuracy. On the RTE task, the best \sys{} model has $0.4\%$ lower accuracy but $1.6\times$ faster runtime. On the WSC task, the best performing \sys{} model obtains a similar one-shot accuracy as OPT-$350$M, but with $1.5\times$ faster runtime. 
\comment{
In the $32$-shot setting (\mojan{Figure~\ref{fig:opt_32shot}} in Appendix~\ref{sec:appdx_opt}), \sys{} models outperform OPT-$350$M on $11$ out of $14$ tasks, delivering accuracy improvements as high as $17.9\%$ for an average $1.2\times$ lower latency. The latency savings can further be pushed up to $1.6\times$ for a lower average accuracy improvement of $4.0\%$ across different tasks.
}



%% file: 5_conclusion.tex
\comment{
\section{Conclusion}\label{sec:conclusion}
We empirically establish a critical insight that there exists a strong correlation between the number of decoder parameters and final model performance for autoregressive Transformers.
Building upon this finding, we develop an efficient on-device search algorithm (LTS) that outputs models on the Pareto-frontier of perplexity versus various hardware metrics, e.g., latency and peak memory utilization. LTS utilizes the decoder parameter count as a training-free and zero-cost proxy for relative ranking of architectures during search. 
Our search can be performed locally on the target (constrained) platform, where hardware performance metrics, e.g., latency, are directly measured. 
We provide large-scale proof-of-concept experiments on the effectiveness of decoder parameter count as a ranking proxy using $2900+$ autoregressive Transformers of varying size, backbones, and two large language datasets.
}

\vspace{-0.3cm}
\section{Limitations and Future Work}\label{sec:conclusion}
\vspace{-0.1cm}
Decoder parameter count provides a simple yet accurate proxy for ranking autoregressive Transformers. This should serve as a strong baseline for future works on Transformer NAS. Our focus is mainly on autoregressive, decoder-only transformers. We therefore, study perplexity as the commonly used metric for language modeling tasks. Nevertheless, recent literature on scaling laws for Transformers suggest a similar correlation between parameter count and task metrics may exist for encoder only (BERT-style) Transformers or encoder-decoder models used in neural machine translation (NMT)~\cite{ghorbani2021scaling}. Additionally, recent findings~\cite{tay2021scale} show specific scaling laws exist between model size and downstream task metrics, e.g., GLUE~\cite{wang2018glue}. Inspired by these observations, we provide preliminary studies that suggest parameter count proxies may be applicable to Transformers in other domains. Detailed investigations of such zero-cost proxies for NAS on heterogeneous BERT-style or NMT models with new performance metrics is an important future avenue of research.

\vspace{-0.2cm}
\section{Acknowledgements}
Professor Farinaz Koushanfar's effort has been in parts supported by the NSF TILOS AI institute, award number CCF-2112665.

%% file: appendix.tex
\setcounter{page}{1}

\section{Preliminaries on Autoregressive Transformers}\label{sec:appdx_prelim}
\noindent\textbf{Perplexity.} Perplexity is a widely used metric for evaluating the performance of autoregressive language models. This metric encapsulates how well the model can predict a word. Formally, perplexity of a language model $M$ is derived using the entropy formula as:
\begin{equation}
    Perplexity(M) = 2^{H(L, M)} = 2^{-\sum_x L(x).log(M(x)))}
\end{equation}
where $L$ represents the ground truth words. As seen, the perplexity is closely tied with the cross-entropy loss of the model, i.e., $H(L, M)$.

\begin{SCfigure}[50][h]
    \centering
    \includegraphics[width=0.45\columnwidth]{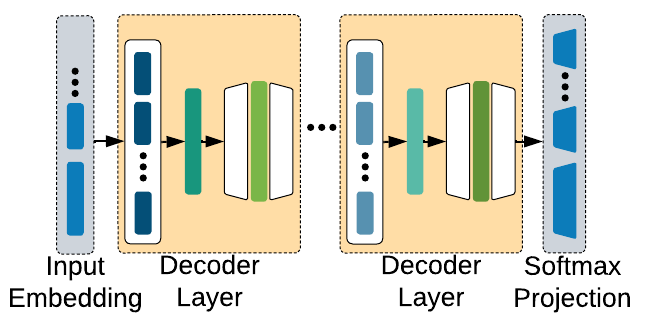}
    \caption{High-level visualization of different components in autoregressive Transformers. Here, the parameters enclosed in the orange blocks are counted as decoder parameters, while the parameters contained in the gray boxes are included in the embedding parameter count.}
    \label{fig:param_count}
    \vspace{-0.5cm}
\end{SCfigure}

\noindent\textbf{Parameter count.} Contemporary autoregressive Transformers consist of three main components, namely, the input embedding layer, hidden layers, and the final (softmax) projection layer. The embedding layer often comprises look-up table-based modules that map the input language tokens to vectors. These vectors then enter a stack of multiple hidden layers a.k.a, the decoder blocks. Each decoder block is made up of an attention layer and a feed-forward network. Once the features are extracted by the stack of decoder blocks, the final prediction is generated by passing through the softmax projection layer. When counting the number of parameters in an autoregressive Transformer, the total parameters enclosed in the hidden layers is dubbed the decoder parameter count or equivalently, the non-embedding parameter count. These parameters are architecture-dependent and do not change based on the underlying tokenization or the vocabulary size. The embedding parameter count, however, accounts for the parameters enclosed in the input embedding layer as well as the final softmax projection layer as they are both closely tied to the word embedding and vocabulary size. We visualize an autoregressive Transformer in Figure~\ref{fig:param_count}, where the orange blocks contain the decoder parameters and grey blocks hold the embedding parameters.

\section{Experimental Setup}\label{sec:appdx_training}
\noindent\textbf{Datasets.} 
We conduct experiments on three datasets, namely, WikiText-103, LM1B, and the Pile. The datasets are tokenized
using word-level and byte-pair encoding for models with Transformer-XL and GPT-2 backbones, respectively.

\noindent\textbf{Training and Evaluation.}
We adopt the open-source code by~\cite{huggingface} and~\cite{transformer_github} to implement the GPT-2 and Transformer-XL backbones, respectively. We further use the source code provided in~\cite{opt_code} to implement the baseline OPT-$350$M and \sys{} models used in zero and one-shot evaluations.
Table~\ref{tab:training_hyperparams} encloses the hyperparameters used for training. 
In this paper, each model is trained separately from scratch. In many scenarios, the user only needs to train one model from the Pareto-frontier, which is selected based on their needs for perplexity, latency, and memory. However, if the users are interested in multiple models, they can either train all models separately or fuse them and train them simultaneously using weight sharing as in~\cite{wang2020hat,zafrir2021prune}. As an example, if the user is interested in two particular models from the Pareto-frontier which have $3$ and $5$ layers, the user can fuse them into a single $5$-layer (super)net and train both models at the same time using weight sharing. The cost of training this supernet is roughly the same as training a $5$-layer model. Therefore, this simple trick can amortize the training cost for Pareto-frontier models.

Throughout the paper, validation perplexity is measured over a sequence length of $192$ and $32$ tokens for WikiText-103 and LM1B datasets, respectively. 
For our zero and one-shot evaluations, we adopt the open-source code by ~\citet{eval-harness}.
Inference latency and peak memory utilization are measured on the target hardware for a sequence length of $192$, averaged over $10$ measurements. The sequence length is increased to $2048$ for latency comparison with the OPT baseline. We utilize PyTorch's native benchmarking interface for measuring the latency and memory utilization of candidate architectures.

\noindent{\textbf{Choice of Training Hyperparameters.}}
For each backbone, dataset, and task, we use the same training setup for all models generated by NAS.
This is the common setting used in the vast majority of NAS papers, including popular benchmarks~\cite{dong2019bench,siems2020bench,zela2021surrogate}, due to the extremely high cost of NAS combined with hyperparameter optimization (HPO).
The setup for our training hyperparameters is based on the evidence provided in prior art in Transformer design~\cite{kaplan2020scaling,raffel2020exploring,ghorbani2021scaling} and NAS~\cite{wang2020hat,zhou2022training,chen2021autoformer}. Specifically, for the range of model sizes studied in this paper, prior work adopts the same batch size (see Table 2.1 in GPT-3~\cite{brown2020language}), which suggests there is no significant benefit in optimizing the batch size per architecture. The original GPT-3 paper~\cite{brown2020language} also adopts the same learning rate scheduler for all models, regardless of their size. Similarly, authors of~\cite{kaplan2020scaling} show that the choice of learning rate scheduler does not have a significant effect on final model performance, which further validates that exploration of the scheduler will not alter the empirical findings in this paper.

Authors of~\cite{kaplan2020scaling} further provide a rule-of-thumb for setting the optimal learning rate (see Equation D.1 of~\cite{kaplan2020scaling}). This rule shows that changes in the optimal learning rate are negligible for the range of model sizes in our search space. We validate this by conducting an experiment that aims to find the optimal learning rate per architecture. We sweep the learning rate $\in[0.0001, 0.001, 0.01, 0.1]$ for $100$ randomly sampled models from the GPT-2 backbone and train them on WikiText-103. The studied models span a wide range of configurations with $2-16$ layers and $2-65$M total parameters. We then pick the optimal learning rate for each architecture, i.e., the one which results in the lowest perplexity. We remeasure the correlation between newly obtained perplexities and the decoder parameter county proxy. Our learning rate optimization experiment results in two important observations: 1) for the vast majority of the architectures ($98\%$), the optimal learning rate is equal to $0.01$, i.e., the value used in all experiments (see Table~\ref{tab:training_hyperparams}), and 2) the ranking of architectures after convergence remains largely unchanged, leading to a correlation of $0.93$ with decoder parameter count, compared to $0.96$ when using the same learning rate for all models. The above evidence suggests that the same training setup can be used for all architectures in the search space, without affecting the results.


\begin{table}[h]
\caption{\sys{} training hyperparameters for different backbones. Here, DO represents dropout layers.}\label{tab:training_hyperparams}
\centering
\resizebox{\columnwidth}{!}{
\begin{tabular}{llcccccccccc}
\hline
Backbone
& Dataset      
& Tokenizer 
& \# Vocab 
& Optim.
& \# Steps 
& Batch size 
& LR     
& Scheduler 
& Warmup 
& DO  
& Attn DO \\ \hline
\multirow{2}{*}{Transformer-XL} 
& WT103 
& Word      
& 267735      
& LAMB~\cite{you2019large}
& 4e4      
& 256       
& 1e-2   
& Cosine    
& 1e3    
& 0.1 & 0.0     \\ \cline{2-12}
& LM1B         
& Word      
& 267735      
& Adam      
& 1e5      
& 224        
& 2.5e-4 & Cosine    
& 2e4    
& 0.0 & 0.0     \\ \hline
\multirow{3}{*}{GPT-2}          
& WT103 
& BPE       
& 50264       
& LAMB~\cite{you2019large}     
& 4e4      
& 256        
& 1e-2   
& Cosine    
& 1e3    
& 0.1 & 0.1     \\ \cline{2-12}
& LM1B           
& BPE       
& 50264       
& LAMB~\cite{you2019large}        
& 1e5      
& 224        
& 2.5e-4 
& Cosine    
& 2e4    
& 0.1 
& 0.1     \\ \cline{2-12}
& Pile
& BPE 
& 50272
& Adam
& 5.48e4
& 256
& 3e-5
& Linear
& 715
& 0.1
& 0.0
 \\ \hline
\end{tabular}}
\end{table}

\noindent \textbf{Search Setup.}
Evolutionary search is performed for $30$ iterations with a population size of $100$; the parent population accounts for $20$ samples out of the total $100$; $40$ mutated samples are generated per iteration from a mutation probability of $0.3$, and $40$ samples are created using crossover.

\input{4_2_zero-cost-NAS}

\section{Analysis on Homogeneous Models}
\label{sec:appdx_homogeneous}
In this section, we evaluate the efficacy of the proposed proxies on the homogeneous search space, i.e., when all decoder layers have the same parameter configuration. In this scenario, the parameters are sampled from the valid ranges in Section~\ref{sec:methodology} to construct one decoder block. This block is then replicated based on the selected n\textsubscript{layer} to create the Transformer architecture. In what follows, we provide experimental results gathered on $100$ randomly sampled Transformer models from the Transformer-XL backbone with homogeneous decoder blocks, trained on WikiText-103.

\begin{figure}[h]
    \centering
    \begin{subfigure}[b]{0.48\textwidth}
     \centering
     \includegraphics[width=\textwidth]{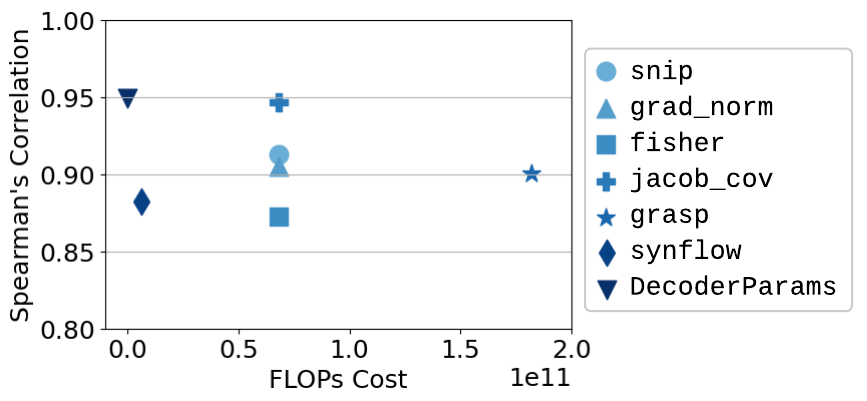}
     \caption{}
    \label{fig:pruning_homogeneous}
    \end{subfigure}
    \begin{subfigure}[b]{0.38\textwidth}
    \centering
    \includegraphics[width=\textwidth]{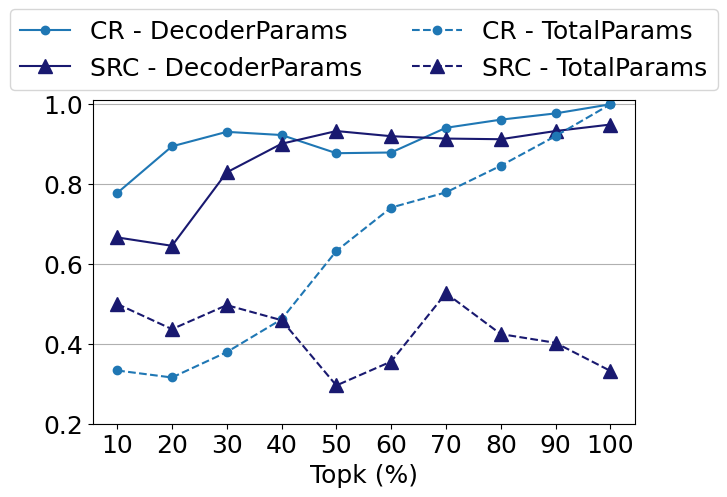}
    \caption{}
    \label{fig:param_count_homogeneous}
    \end{subfigure}
    \vspace{-0.3cm}
\caption{Experiments conducted on $100$ randomly sampled Transformers with homogeneous decoder blocks, trained on WikiText-103. (a)~SRC between the ranking obtained from low-cost proxies and the ground truth ranking after full training. The decoder parameter count obtains the best SRC with zero cost. (b)~Performance of parameter count proxies. The decoder parameter count provides a very accurate ranking proxy with an SRC of $0.95$ over all models.}
\vspace{-0.5cm}
\end{figure}

\noindent $\blacktriangleright$ \textbf{Low-cost Proxies.} Figure~\ref{fig:pruning_homogeneous} demonstrates the SRC between various low-cost methods and the validation perplexity after full training. On the horizontal axis, we report the total computation required for each proxy in terms of FLOPs. Commensurate with the findings on the heterogeneous models, we observe a strong correlation between the low-cost proxies and validation perplexity, with the decoder parameter count outperforming other proxies. Note that we omit the \relulogdet method from Figure~\ref{fig:pruning_homogeneous} as it provides a low SRC of $0.42$ due to heavy reliance on ReLU activations.

\noindent $\blacktriangleright$ \textbf{Parameter Count.} As seen in Figure~\ref{fig:param_count_homogeneous}, the total parameter count has a low SRC with the validation perplexity while the decoder parameter count provides an accurate proxy with an SRC of $0.95$ over all architectures. These findings on the homogeneous search space are well-aligned with the observations in the heterogeneous space.

\vspace{-0.2cm}
\section{How Does Model Topology Affect the Training-free Proxies?}\label{sec:appdx_scaling}

\begin{figure}[h]
    \centering
    \begin{subfigure}[b]{0.32\textwidth}
     \centering
    \includegraphics[width=\columnwidth]{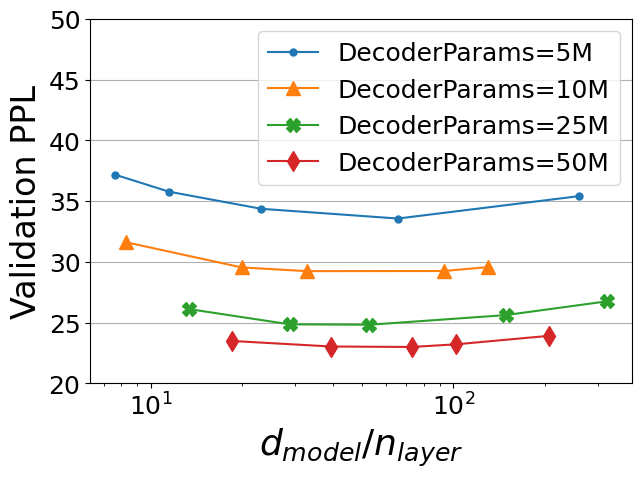}
    \caption{}\label{fig:scaling_exps_a}
    \end{subfigure}
    \begin{subfigure}[b]{0.32\textwidth}
     \centering
     \includegraphics[width=\columnwidth]{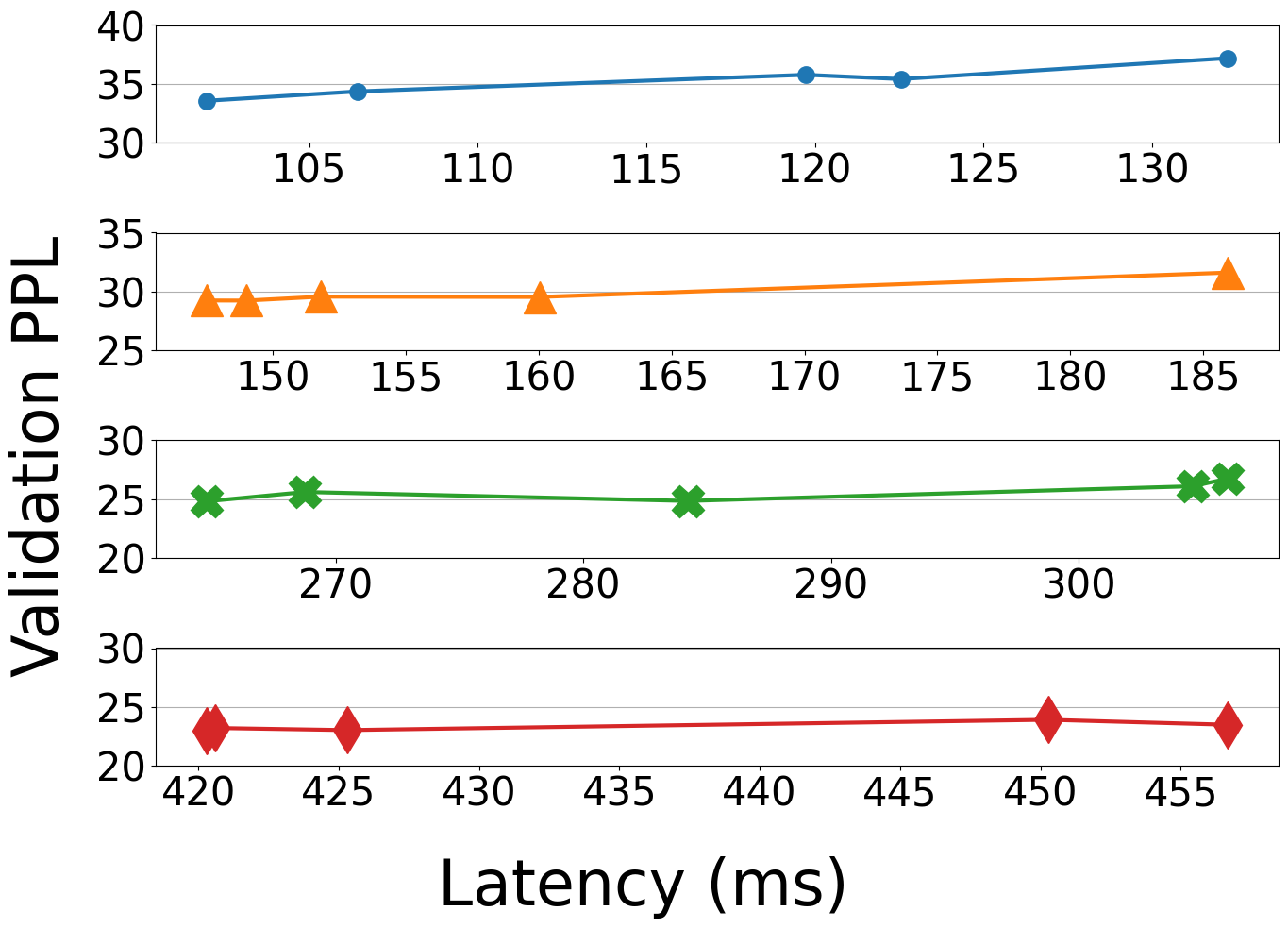}
    \caption{}\label{fig:scaling_exps_b}
    \end{subfigure}
    \begin{subfigure}[b]{0.32\textwidth}
     \centering
     \includegraphics[width=\columnwidth]{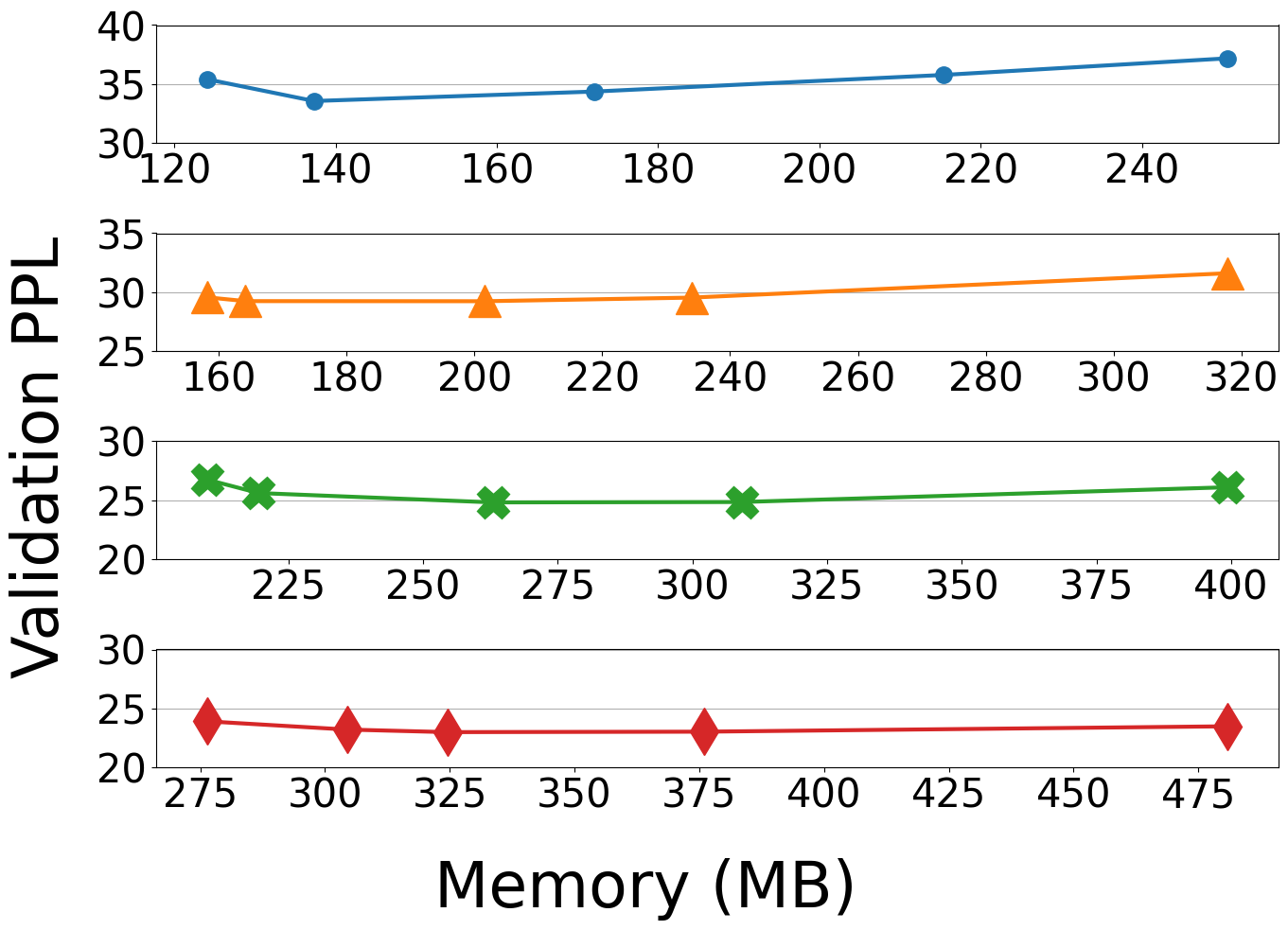}
    \caption{}\label{fig:scaling_exps_c}
    \end{subfigure}
    \vspace{-0.2cm}
    \caption{Validation perplexity after full training versus (a) the width-to-depth aspect ratio, (b) latency, and (c) peak memory utilization. Models are randomly generated from the Transformer-XL backbone and trained on WikiText-103. For a given decoder parameter count, we observe low variation in perplexity across different models, regardless of their topology. The topology, however, significantly affects the latency and peak memory utilization for models with the same perplexity.}
    \label{fig:scaling_exps}
\end{figure}

Figure~\ref{fig:scaling_exps_a} shows the validation perplexity versus the aspect ratio of random architectures sampled from the Transformer-XL backbone and trained on WikiText-103. Here, the models span wide, shallow topologies (e.g., d\textsubscript{model}=$1024$, n\textsubscript{layer}=$3$) to narrow, deep topologies (e.g., d\textsubscript{model}=$128$, n\textsubscript{layer}=$35$).
The maximum change in the validation perplexity for a given decoder parameter count is $<7\%$ for a wide range of aspect ratios $\in[8,323]$. Nevertheless, for the same decoder parameter count budget, the latency and peak memory utilization vary by $1.3\times$ and $2.0\times$ as shown in Figures~\ref{fig:scaling_exps_b} and~\ref{fig:scaling_exps_c}.

For deeper architectures (more than $40$ layers) with the Transformer-XL backbone, we observe an increase in the validation perplexity, which results in a deviation from the pattern in Figure~\ref{fig:scaling_exps}a. This observation is associated with the inherent difficulty in training deeper architectures, which can be mitigated with the proposed techniques in the literature~\cite{wang2022deepnet}. Nevertheless, such deep models have a high latency, which makes them unsuitable for lightweight inference. For hardware-aware and efficient Transformer NAS, our search space contains architectures with less than $16$ layers. In this scenario, the decoder parameter count proxy holds a very high correlation with validation perplexity, regardless of the architecture topology as shown in Figure~\ref{fig:scaling_exps_a}. 

\vspace{-0.2cm}
\section{3D Pareto Visualization}\label{sec:appdx_3dpareto}
\begin{SCfigure}[10][t]
    \centering
    \includegraphics[width=0.6\columnwidth]{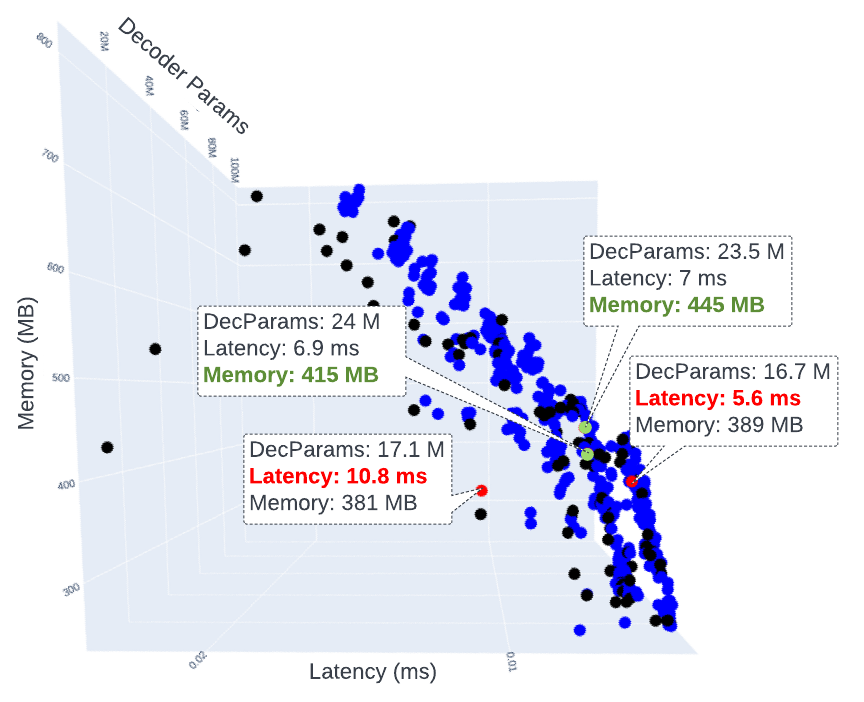}
    \caption{3D visualization of our multi-objective NAS for the GPT-2 backbone on TITAN Xp GPU. Architectures with similar memory and decoder parameter count can result in drastically different runtimes (up to $2\times$ difference). Similarly, architectures with similar decoder parameter count and latency may have different peak memory utilization. Therefore, it is important to perform multi-objective NAS where several hardware characteristics are simultaneously taken into account when extracting the Pareto-frontier.}
    \label{fig:3d_pareto}
    \vspace{-0.5cm}
\end{SCfigure}

Figure~\ref{fig:3d_pareto} visualizes the $3$-dimensional Pareto obtained during search on the GPT-2 backbone. Here, the black and blue points denote regular and Pareto-frontier architectures, respectively. The pair of red dots are architectures which match in both memory and decoder parameter count ($\sim$ perplexity). However, as shown, their latency differs by $2\times$. The pair of green points correspond to models with the same decoder parameter count ($\sim$ perplexity) and latency, while the memory still differs by $30$MB, which is non-negligible for memory-constrained application. In a $2$-objective Pareto-frontier search of perplexity versus memory (or latency), each pair of red (or green) dots will result in similar evaluations. While in reality, they have very different characteristics in terms of the overlooked metric. This experiment validates the need for multi-objective Pareto-frontier search, which simultaneously takes into account multiple hardware performance metrics. 

\begin{figure*}[h]
    \centering
    \includegraphics[width=\textwidth]{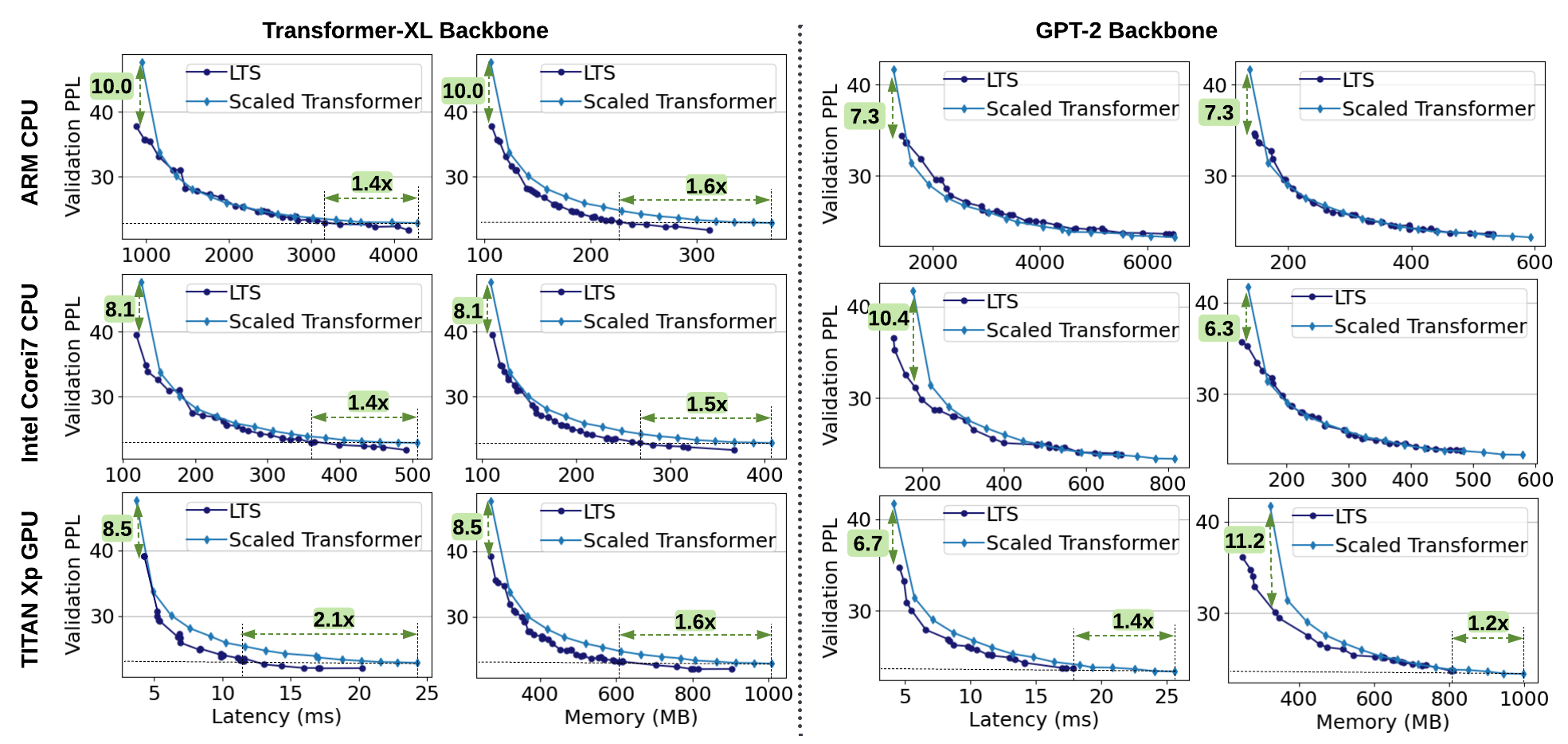}
    \vspace{-0.4cm}
    \caption{2D visualization of the perplexity versus latency and memory Pareto-frontier found by \sys{} and scaled backbone models with varying number of layers. All models are trained on the WikiText-103 dataset. The architectural parameters for all models are enclosed in Appendix~\ref{sec:appdx_archs}.}
    \label{fig:nas_results_wt103}
\vspace{-0.3cm}
\end{figure*}

\section{\sys{} Pareto-frontier on WikiText-103}\label{sec:comp_wt103}
\vspace{-0.2cm}
We compare the Pareto-frontier architectures found by \sys{} with the baseline after full training on the WikiText-103 dataset in Figure~\ref{fig:nas_results_wt103}. Commensurate with the findings on the LM1B dataset, the NAS-generated models outperform the baselines in at least one of the three metrics, i.e., perplexity, latency, and peak memory utilization. We note that the gap between the baseline models and those obtained from NAS is larger when training on the LM1B dataset. This is due to the challenging nature of LM1B, which exceeds the WikiText-103 dataset size by $\sim 10\times$. Thus, it is harder for hand-crafted baseline models to compete with the optimized \sys{} architectures on LM1B.

On the Transformer-XL backbone, the models on \sys{} Pareto-frontier for the ARM CPU have, on average, $3.8\%$ faster runtime and $20.7\%$ less memory under the same validation perplexity budget. On the Corei7, the runtime and memory savings increase to $13.2\%$ and $19.6\%$, respectively, while matching the baseline perplexity. 
We achieve our highest benefits on TITAN Xp GPU where \sys{} Pareto-frontier models have, on average, $31.8\%$ lower latency and $21.5\%$ lower memory utilization. Notably, the validation perplexity of the baseline $16$-layer Transformer-XL base can be achieved with a lightweight model with $2.1\times$ lower latency while consuming $1.6\times$ less memory at runtime.

On the GPT-2 backbone, \sys{} achieves $6.3-11.2$ lower perplexity in the low-latency-and-memory regime. As we transition to larger models and higher latency, our results show that the GPT-2 architecture is nearly optimal
on WikiText-103 when performing inference on a CPU. 
The benefits are more significant when targeting a GPU; For any given perplexity achieved by the baseline, \sys{} Pareto-frontier on TITAN Xp delivers, on average, $9.0\%$ lower latency and $4.5\%$ lower memory. Therefore, the perplexity and memory of the baseline $16$-layer GPT-2 can be achieved by a new model that runs $1.4\times$ faster and consumes $1.2\times$ less memory on TITAN Xp.

\section{Zero and One-Shot Evaluation of \sys{} Models}\label{sec:appdx_opt}


\begin{figure}[h]
     \centering
     \begin{subfigure}[b]{0.48\textwidth}
         \centering
         \includegraphics[width=\textwidth]{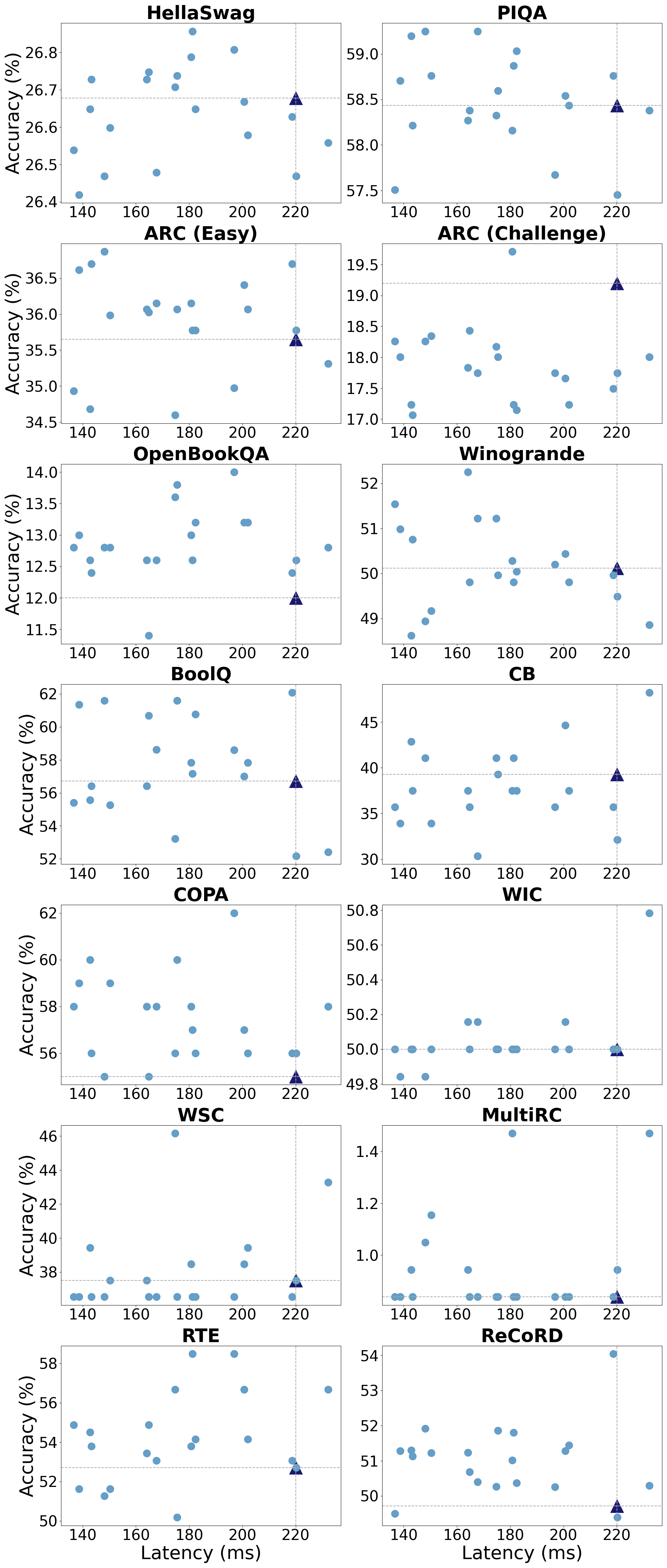}
         \caption{zero-shot}
         \label{fig:appdx_opt_0shot}
     \end{subfigure}
     \hfill
     \begin{subfigure}[b]{0.48\textwidth}
         \centering
         \includegraphics[width=\textwidth]{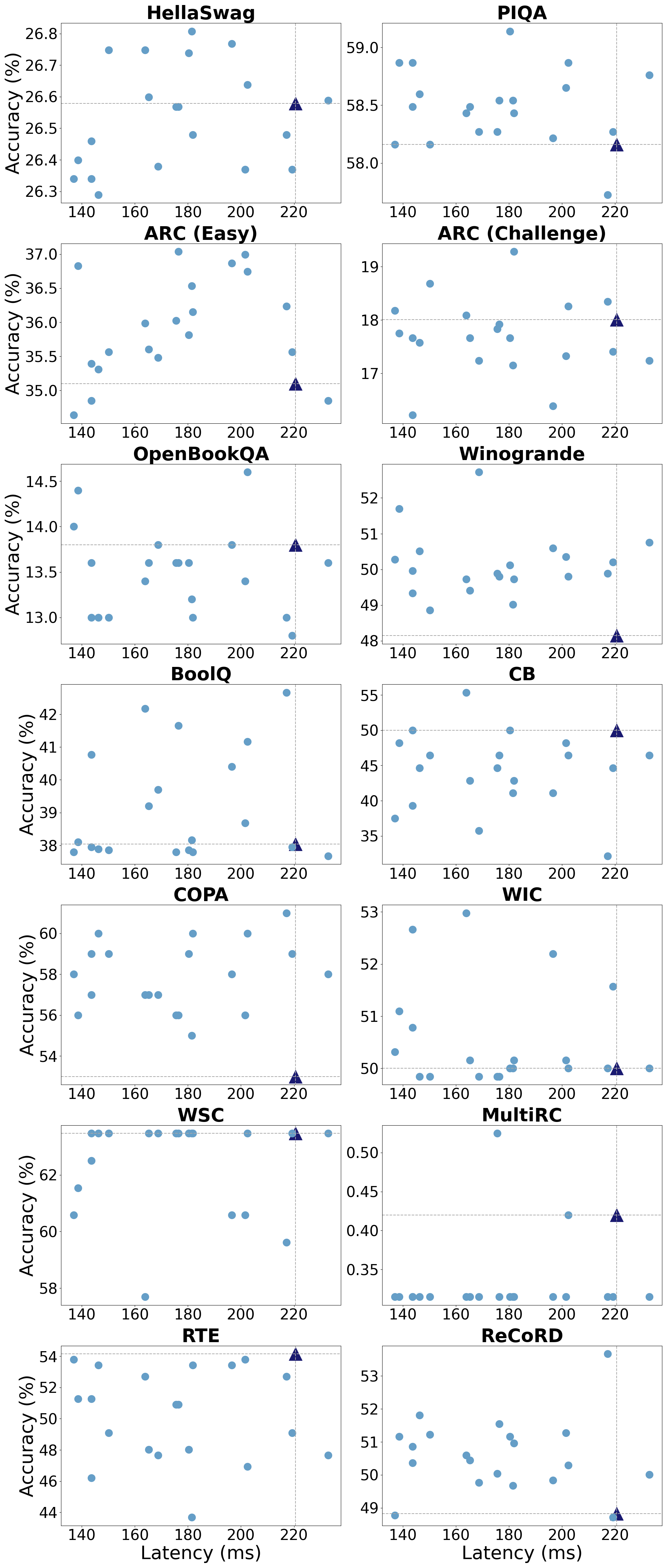}
         \caption{one-shot}
         \label{fig:appdx_opt_1shot}
     \end{subfigure}
        \caption{
        \sys{} Pareto-frontier models (dots) achieve a higher zero and one-shot accuracy with lower latency compared to the hand-designed OPT-$350$M model (triangle).
        Latency is measured on an A6000 NVIDIA GPU. Architectural parameters for all models shown here are detailed in Appendix~\ref{sec:appdx_archs}.}
        \label{fig:appdx_opt}
\end{figure}

For this experiment, we design our search space to cover models with a similar parameter count budget as the OPT-$350$M model. To this end, we search over the following values for the architectural parameters: n\textsubscript{layer}$\in \{3,\dots, 29|1\}$, d\textsubscript{model}$\in \{512,\dots,1472|64\}$, d\textsubscript{inner}$\in \{512,\dots,6080|64\}$, and n\textsubscript{head}$\in \{2,4,8,16\}$. To directly compare with OPT, we use a generic, non-adaptive embedding layer for our models. Therefore, the search space does not include the $k$ factor and d\textsubscript{embed}=d\textsubscript{model}.

Figures~\ref{fig:appdx_opt_0shot} and~\ref{fig:appdx_opt_1shot} show the per-task zero and one-shot performance of \sys{} models and OPT-$350$M.
Please refer to Section~\ref{sec:opt} of the main paper for a summarization of the results in these figures.

\section{Architecture Details}\label{sec:appdx_archs}
Tables~\ref{tab:archs_transxl},~\ref{tab:archs_transxl2},~\ref{tab:archs_gpt},~\ref{tab:archs_gpt2} enclose the architecture parameters for the baseline and NAS-generated models in Figures~\ref{fig:nas_results} and~\ref{fig:nas_results_wt103} for Transformer-XL and GPT-2 backbones. Table~\ref{tab:archs_opt} further holds the architecture details of models used in our zero and one-shot evaluations of Figures~\ref{fig:avg_acc_opt} and~\ref{fig:appdx_opt}.
For each target hardware, the rows of the table are ordered based on increasing decoder parameter count (decreasing validation perplexity). For all models, d\textsubscript{head}=d\textsubscript{model}/n\textsubscript{head} and d\textsubscript{embed}=d\textsubscript{model}. For models in Tables~\ref{tab:archs_transxl},~\ref{tab:archs_transxl2},~\ref{tab:archs_gpt},~\ref{tab:archs_gpt2}, the adaptive input embedding factor is set to $k=4$. The models in Table~\ref{tab:archs_opt}, however, use the generic, non-adaptive, input embedding ($k=1$) following the original OPT architecture~\cite{zhang2022opt}.


\section{Transformers in other Domains}\label{sec:appdx_domains}
In what follows, we perform preliminary experiments on Transformers used on other domains to investigate the applicability of parameter-based proxies for ranking.

\begin{SCfigure}[50][h]
\vspace{-0.3cm}
    \centering
    \includegraphics[width=0.48\textwidth]{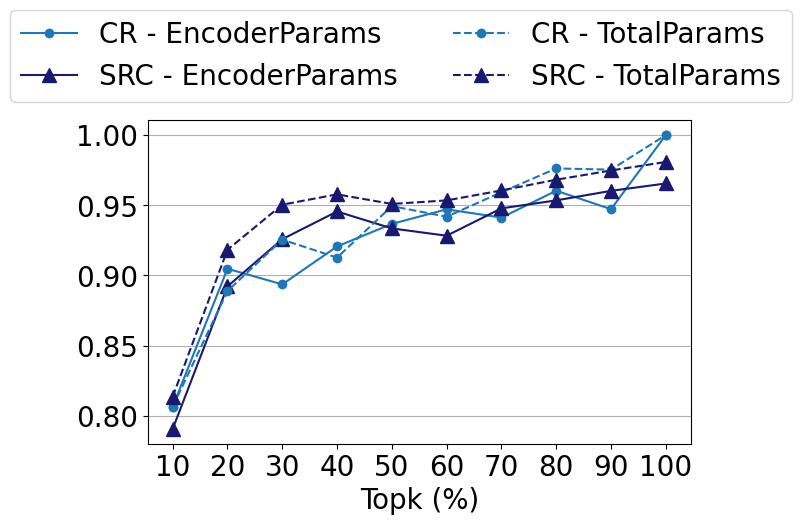}
    \caption{Performance of parameter count proxies on $300$ randomly sampled models from the BERT backbone, trained on WikiText-103. Both encoder and total parameter counts provide a
very accurate ranking proxy with an SRC of 0.96 and 0.98 over all models, respectively.}
    \label{fig:bert}
\end{SCfigure}

\noindent\textbf{Encoder-only Transformers.} BERT~\cite{devlin2019bert} is a widely popular Transformer composed of encoder blocks, which is used in a variety of tasks, e.g., question answering and language inference. The main difference between BERT and the Transformers studied in this paper is the usage of bidirectional versus causal attention. Specifically, the encoder blocks in BERT are trained to compute attention between each input token and all surrounding tokens. In autoregressive models, however, attention is only computed for tokens appearing prior to the current token. BERT is trained with a mixture of masked language modeling and next sentence prediction objectives to ensure applicability to language modeling as well as downstream language understanding tasks. We use the architectural parameters described in Section~\ref{sec:methodology} to construct the search space and randomly sample $300$ models from the BERT backbone. We then train all models on WikiText-103 for 40K steps following the training setup provided in the original BERT paper~\cite{devlin2019bert} for the batch size, sequence length, optimizer, learning rate, vocabulary size, and tokenizer. Figure~\ref{fig:bert} demonstrates the CR and SRC of encoder parameter count and test perplexity measured on various top$k\%$ performing BERT models. As seen, both the encoder and total parameter count provide a highly accurate proxy for test perplexity of BERT, achieving an SRC of $0.96$ and $0.98$, respectively. This trend suggests that parameter-based proxies for NAS can be applicable to encoder-only search spaces as well.

\noindent\textbf{Encoder-Decoder Transformers.} Transformers in this domain comprise both encoder and decoder layers with bidirectional and causal attention computation. This unique structure makes these models suitable for sequence-to-sequence tasks such as Neural Machine Translation (NMT). Recent work~\cite{ghorbani2021scaling} shows that the performance of encoder-decoder Transformers also follows a scaling law with model size. This power-law behavior between model size and performance can be leveraged to develop training-free proxies for ranking these architectures during search. We test our hypothesis by performing experiments on the open-source NMT benchmark by~\cite{zhang2020reproducible,nmt_github} which consists of $2000$ Transformers trained on various language pairs. The pre-trained Transformers in this benchmark have homogeneous layers, i.e., the architectural parameters are the same for all layers and identical for the encoder and the decoder. In addition to architectural parameters, the search space for this benchmark also includes various BPE tokenization and learning rates. We, therefore, pre-process the benchmark by gathering all instances of Transformers for a fixed BPE. Then for each given architecture, we keep the results corresponding to the best-performing learning rate.

\begin{figure}[h]
    \centering
    \includegraphics[width=\textwidth]{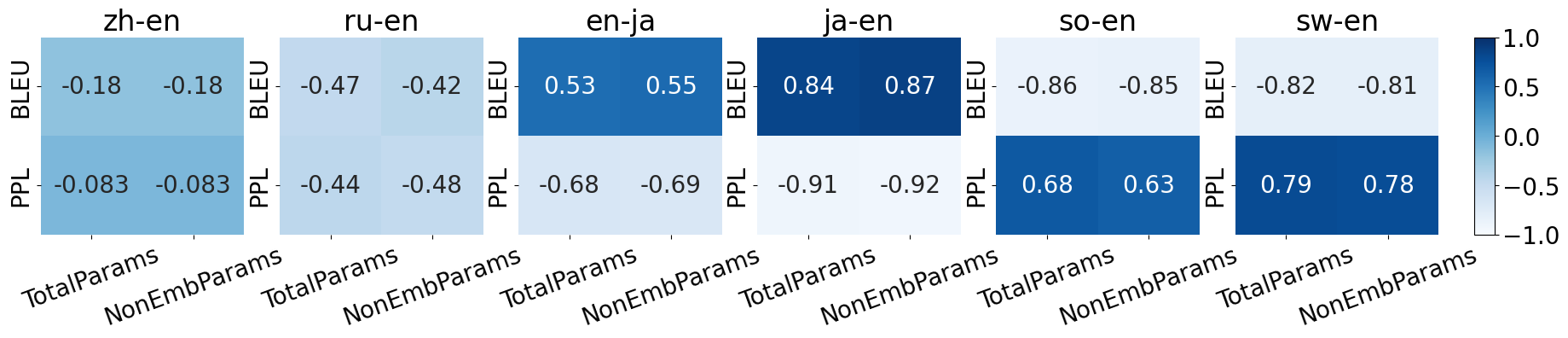}
    \caption{SRC between parameter count proxies and performance metrics, i.e., BLEU score and perplexity (PPL) for translation between various language pairs. The ``NonEmbParams'' label denotes the parameters enclosed in the encoder and decoder blocks while the ``TotalParams'' label corresponds to the total parameter count including those in the embedding layers.  Here, darker versus lighter colors show a high positive and negative correlation, respectively.}
    \label{fig:nmt}
\end{figure}

Figure~\ref{fig:nmt} shows a heatmap of the SRC between parameter count proxies and perplexity as well as the BLEU score. As seen, the ranking performance of total parameter count versus non-embedding parameter count, i.e., parameters enclosed in the encoder and decoder blocks, is largely similar. On certain tasks, e.g., `ja-en', `so-en', and `sw-en' the parameter count proxies perform quite well, achieving a high SRC with both the BLEU score and perplexity. Interestingly, on `so-en' and `sw-en', the parameter count and performance are inversely correlated, which may be due to the limited training data for these language pairs which gives smaller models a leading advantage over larger architectures. While these preliminary results show promise for parameter-based proxies in NAS for NMT, several aspects require further investigation, e.g., the effect of architectural heterogeneity and dataset size on the performance of these proxies. Studying these aspects may perhaps lead to a new formulation of training-free proxies for NMT and are out of scope for this paper.

\vspace{-0.5cm}
\section{Ethics Statement and Broader Impact}\label{sec:ethics}
We provide an extremely lightweight method for NAS on autoregressive Transformers. Our work is likely to increase the adoption of NAS in the NLP domain, providing several prevalent benefits:

Firstly, more widespread adoption of automated techniques, e.g., NAS eliminates the need for laborious trials and error for manual design of Transformer architectures, freeing up hundreds of hours of man-power as well as computational resources. Secondly, automating architecture design can trigger the generation of new models with superior performance, which benefits the ever-growing applications of NLP in the everyday life. Finally, by making the search algorithm efficient, we ensure it can be accessible to the general scientific public without need for any expensive mode training, thereby minimizing the unwanted byproducts of the Deep Learning era such as the carbon footprint, and power consumption.
While the benefits of automation in NLP are plenty, it can lead to potential side effects that have not been yet fully unveiled.
Since our work advances the use of NAS in the NLP design pipeline, there is need for scrutiny of the models which have been automatically designed with respect to aspects such as bias, misinformation, and nefarious activity, to name a few.



\begin{table}[h]
\centering
\caption{Detailed architectural parameters for all models in Figure~\ref{fig:avg_acc_opt} with GPT-2 backbone.}\label{tab:archs_opt}
\resizebox{0.53\textwidth}{!}{
\begin{tabular}{cccccc} \hline
& n\textsubscript{layer} 
& d\textsubscript{model} 
& n\textsubscript{head} 
& d\textsubscript{inner}
& DecoderParams (M) \\ \hline 
baseline (OPT-$350$M) & 24 & 1024 & 16 & 4096 & 304.4 \\ \hline
M1 &26 &1024 &16 &2816 &261.4 \\ 
M2 &15 &1280 &16 &4480 &273.2 \\ 
M3 &24 &1280 &8 &1856 &274.3 \\ 
M4 &16 &1344 &8 &3840 &283.8 \\ 
M5 &14 &1344 &8 &4800 &284.8 \\ 
M6 &20 &1216 &4 &3456 &289.2 \\ 
M7 &16 &1344 &16 &4096 &294.8 \\ 
M8 &28 &1344 &8 &1344 &306.6 \\ 
M9 &28 &1088 &8 &2816 &306.7 \\ 
M10 &26 &1152 &16 &2816 &309.4 \\ 
M11 &25 &832 &2 &5760 &310.9 \\ 
M12 &20 &1280 &16 &3456 &310.9 \\ 
M13 &19 &1280 &8 &3840 &314.2 \\ 
M14 &26 &1152 &4 &3008 &320.9 \\ 
M15 &19 &1472 &8 &2816 &325.5 \\ 
M16 &13 &1472 &4 &5568 &329.0 \\ 
M17 &14 &1480 &2 &5824 &367.3 \\ 
M18 &20 &1152 &8 &5760 &374.3 \\ 
M19 &26 &1024 &4 &5696 &414.8 \\ 
M20 &25 &1408 &8 &3136 &422.3 \\ 
\hline
\end{tabular}}
\end{table}

\begin{table}[h]
\caption{Detailed architectural parameters for all models in Figure~\ref{fig:nas_results} with Transformer-XL backbone.}\label{tab:archs_transxl}
\centering
\resizebox{\textwidth}{!}{
\begin{tabular}{lcccccc} \hline
\multicolumn{2}{c}{}
& n\textsubscript{layer} 
& d\textsubscript{model} 
& n\textsubscript{head} 
& d\textsubscript{inner}
& DecoderParams (M) \\ \hline 
\multicolumn{2}{l}{baseline} & $\in$[1,16] & 512 & 8 & 2048 & - \\ \hline
\multirow{39}{*}{\rotatebox{90}{ARM}}&M1 &2 &512 &[2, 2] &[1216, 1280] &5.2 \\ 
&M2 &3 &320 &[2, 4, 2] &[1472, 2368, 3392] &6.2 \\ 
&M3 &2 &512 &[2, 2] &[2560, 2176] &7.5 \\ 
&M4 &2 &512 &[2, 2] &[3904, 1792] &8.5 \\ 
&M5 &2 &640 &[2, 2] &[3520, 3456] &13.0 \\ 
&M6 &2 &704 &[8, 2] &[3904, 3968] &16.1 \\ 
&M7 &2 &832 &[2, 2] &[3264, 3968] &19.0 \\ 
&M8 &2 &960 &[2, 2] &[3648, 3968] &23.9 \\ 
&M9 &2 &960 &[2, 2] &[3904, 3968] &24.4 \\ 
&M10 &3 &960 &[2, 2, 2] &[1856, 2368, 3392] &28.5 \\ 
&M11 &3 &832 &[2, 2, 2] &[3904, 3968, 3008] &28.5 \\ 
&M12 &3 &960 &[2, 4, 2] &[3328, 2368, 3200] &30.9 \\ 
&M13 &3 &960 &[4, 2, 2] &[3648, 3584, 3584] &34.6 \\ 
&M14 &3 &960 &[2, 2, 2] &[3904, 3584, 3456] &34.9 \\ 
&M15 &3 &960 &[2, 2, 8] &[4032, 3968, 3904] &36.7 \\ 
&M16 &4 &896 &[4, 2, 8, 2] &[3904, 3008, 3520, 3584] &41.2 \\ 
&M17 &4 &960 &[8, 8, 8, 4] &[4032, 3968, 2880, 3200] &45.5 \\ 
&M18 &4 &960 &[2, 2, 2, 2] &[3840, 3904, 3520, 3072] &46.0 \\ 
&M19 &4 &960 &[2, 2, 2, 2] &[4032, 3648, 3136, 4032] &47.0 \\ 
&M20 &4 &960 &[8, 2, 4, 8] &[4032, 3584, 3840, 3584] &47.4 \\ 
&M21 &4 &960 &[2, 2, 4, 2] &[3904, 3968, 3840, 3584] &47.8 \\ 
&M22 &5 &960 &[2, 2, 2, 2, 2] &[3904, 3968, 3264, 3456, 3200] &57.3 \\ 
&M23 &5 &960 &[2, 2, 2, 8, 2] &[3904, 3648, 3136, 3648, 3840] &58.0 \\ 
&M24 &6 &960 &[2, 2, 2, 2, 2, 8] &[3328, 2624, 3392, 2944, 3008, 3904] &64.6 \\ 
&M25 &6 &960 &[2, 2, 4, 2, 8, 8] &[3584, 2624, 3392, 3968, 3008, 3328] &65.9 \\ 
&M26 &6 &960 &[2, 4, 2, 2, 2, 2] &[2112, 3840, 3328, 3264, 3968, 3648] &66.4 \\ 
&M27 &6 &960 &[2, 4, 2, 2, 8, 2] &[3904, 3008, 3392, 3648, 3392, 3584] &67.9 \\ 
&M28 &6 &960 &[2, 2, 2, 2, 2, 4] &[3968, 3968, 3456, 3456, 3776, 2432] &68.1 \\ 
&M29 &6 &960 &[2, 4, 8, 4, 2, 8] &[3904, 3008, 3392, 3200, 3968, 3904] &68.8 \\ 
&M30 &6 &960 &[8, 8, 2, 4, 2, 4] &[3904, 3648, 3136, 3648, 3200, 3840] &68.8 \\ 
&M31 &6 &960 &[8, 4, 8, 4, 2, 8] &[3904, 3648, 3392, 3200, 3968, 3840] &69.9 \\ 
&M32 &8 &896 &[4, 2, 2, 4, 4, 2, 4, 8] &[3584, 3968, 3392, 3904, 2240, 1856, 2560, 3264] &76.6 \\ 
&M33 &8 &896 &[4, 2, 2, 2, 4, 4, 4, 2] &[3584, 3584, 3520, 2368, 2752, 4032, 3520, 3264] &79.9 \\ 
&M34 &9 &896 &[4, 2, 4, 4, 8, 2, 8, 8, 2] &[3840, 3136, 3520, 2880, 3200, 3008, 3328, 2560, 3136] &87.5 \\ 
&M35 &8 &960 &[2, 4, 4, 4, 4, 8, 2, 2] &[3968, 3584, 3520, 3072, 3968, 4032, 1856, 3712] &90.2 \\ 
&M36 &12 &832 &[2, 4, 4, 2, 2, 8, 8, 8, 4, 4, 2, 8] &[3136, 2112, 2112, 2368, 2752, 2432, 2432, 2176, 3456, 3712, 2880, 3712] &97.0 \\ 
&M37 &9 &960 &[4, 4, 8, 2, 2, 2, 8, 8, 2] &[2112, 3008, 3520, 3648, 3968, 4032, 1984, 3200, 3520] &97.2 \\ 
&M38 &9 &960 &[8, 2, 4, 2, 8, 8, 8, 2, 2] &[3968, 3008, 3520, 3200, 3200, 4032, 1984, 2816, 3520] &97.7 \\ 
&M39 &12 &832 &[4, 4, 4, 2, 2, 8, 4, 8, 2, 8, 2, 8] &[3136, 3968, 2112, 2368, 3072, 2240, 2624, 2112, 3456, 3072, 2880, 3264] &98.7 \\ 
\hline
\multirow{35}{*}{\rotatebox{90}{Corei7}}&M1 &2 &384 &[2, 2] &[896, 2816] &4.3 \\ 
&M2 &2 &576 &[2, 2] &[1792, 2816] &8.6 \\ 
&M3 &2 &576 &[2, 2] &[1408, 3776] &9.3 \\ 
&M4 &2 &832 &[2, 2] &[1728, 1536] &12.4 \\ 
&M5 &2 &768 &[2, 2] &[3776, 1920] &14.7 \\ 
&M6 &2 &768 &[2, 2] &[2112, 3584] &14.7 \\ 
&M7 &2 &832 &[2, 2] &[3776, 3392] &18.9 \\ 
&M8 &2 &832 &[2, 2] &[3968, 3584] &19.5 \\ 
&M9 &2 &960 &[2, 4] &[1984, 3840] &20.4 \\ 
&M10 &2 &960 &[8, 8] &[3968, 3584] &23.7 \\ 
&M11 &2 &960 &[2, 2] &[3904, 3904] &24.2 \\ 
&M12 &3 &896 &[2, 2, 2] &[2304, 3904, 3904] &30.2 \\ 
&M13 &3 &960 &[2, 2, 4] &[2176, 3840, 2880] &30.9 \\ 
&M14 &3 &960 &[2, 2, 4] &[3776, 2880, 3904] &34.1 \\ 
&M15 &3 &960 &[2, 8, 2] &[3840, 3840, 3904] &36.1 \\ 
&M16 &3 &960 &[2, 8, 8] &[3904, 3840, 3904] &36.2 \\ 
&M17 &3 &960 &[2, 2, 8] &[3968, 3904, 3904] &36.5 \\ 
&M18 &4 &960 &[2, 4, 2, 2] &[3904, 2112, 4032, 3584] &44.6 \\ 
&M19 &4 &960 &[2, 2, 2, 4] &[2112, 3840, 3904, 3904] &44.9 \\ 
&M20 &4 &960 &[2, 4, 8, 4] &[3776, 3392, 3520, 3904] &46.5 \\ 
&M21 &4 &960 &[2, 2, 2, 4] &[3904, 3776, 3904, 3904] &48.2 \\ 
&M22 &5 &960 &[2, 2, 2, 2, 2] &[3776, 1984, 3904, 3904, 3456] &55.8 \\ 
&M23 &5 &960 &[2, 4, 2, 4, 2] &[3968, 3584, 3520, 3904, 3200] &58.0 \\ 
&M24 &5 &960 &[2, 4, 4, 4, 2] &[3776, 3840, 3904, 3904, 3968] &60.3 \\ 
&M25 &6 &960 &[2, 4, 4, 2, 2, 4] &[3776, 3840, 3904, 3904, 3008, 2304] &67.5 \\ 
&M26 &6 &960 &[2, 4, 2, 4, 2, 4] &[3776, 2112, 4032, 3584, 3200, 4032] &67.5 \\ 
&M27 &6 &960 &[2, 4, 2, 4, 4, 4] &[3776, 3840, 3904, 4032, 3648, 2432] &69.2 \\ 
&M28 &6 &960 &[4, 2, 8, 4, 2, 2] &[3840, 3712, 3520, 4032, 3200, 4032] &70.6 \\ 
&M29 &7 &960 &[2, 2, 8, 4, 2, 2, 4] &[3776, 3840, 3904, 1856, 3072, 3648, 4032] &78.7 \\ 
&M30 &8 &960 &[2, 2, 2, 4, 2, 4, 8, 2] &[3392, 1792, 3904, 3904, 3200, 2432, 1792, 2496] &80.9 \\ 
&M31 &8 &960 &[2, 2, 4, 4, 2, 4, 8, 2] &[3776, 3008, 4032, 3904, 3520, 3136, 1984, 3648] &88.8 \\ 
&M32 &8 &960 &[8, 2, 2, 4, 8, 4, 4, 8] &[3776, 3008, 3904, 3904, 2176, 4032, 4032, 3648] &91.6 \\ 
&M33 &13 &768 &[2, 8, 2, 4, 2, 2, 4, 2, 2, 8, 8, 8, 4] &[3776, 2112, 1600, 3904, 3840, 2880, 2304, 3200, 2048, 2944, 2816, 3328, 3968] &97.9 \\ 
&M34 &9 &960 &[4, 2, 4, 4, 4, 4, 8, 8, 2] &[3840, 3136, 3520, 4032, 3200, 4032, 3648, 2112, 2368] &98.9 \\ 
&M35 &9 &960 &[8, 2, 8, 8, 2, 4, 8, 2, 2] &[3520, 3008, 2880, 4032, 3200, 2432, 4032, 3904, 3136] &99.4 \\  
\hline
\end{tabular}}
\end{table}

\begin{table}[h]
\caption{Detailed architectural parameters for all models in Figure~\ref{fig:nas_results} with Transformer-XL backbone.}\label{tab:archs_transxl2}
\centering
\resizebox{\textwidth}{!}{
\begin{tabular}{lcccccc} \hline
\multicolumn{2}{c}{}
& n\textsubscript{layer} 
& d\textsubscript{model} 
& n\textsubscript{head} 
& d\textsubscript{inner}
& DecoderParams (M) \\ \hline 
\multicolumn{2}{l}{baseline} & $\in$[1,16] & 512 & 8 & 2048 & - \\ \hline
\multirow{40}{*}{\rotatebox{90}{TITAN Xp}}&M1 &2 &384 &[2, 2] &[1152, 2432] &4.2 \\ 
&M2 &2 &448 &[8, 2] &[2944, 3008] &7.4 \\ 
&M3 &2 &576 &[2, 2] &[2048, 1728] &7.7 \\ 
&M4 &2 &512 &[2, 2] &[2368, 3072] &8.2 \\ 
&M5 &2 &832 &[8, 2] &[3264, 3072] &17.5 \\ 
&M6 &2 &768 &[2, 2] &[3968, 4032] &18.2 \\ 
&M7 &2 &896 &[8, 4] &[4032, 2880] &20.4 \\ 
&M8 &2 &960 &[4, 8] &[3968, 3008] &22.6 \\ 
&M9 &2 &960 &[4, 8] &[3968, 3648] &23.9 \\ 
&M10 &2 &960 &[2, 2] &[3840, 3968] &24.2 \\ 
&M11 &3 &896 &[8, 4, 8] &[4032, 2112, 3392] &29.2 \\ 
&M12 &3 &896 &[2, 2, 2] &[3840, 2880, 3840] &31.0 \\ 
&M13 &3 &960 &[2, 2, 2] &[3584, 3072, 2624] &31.7 \\ 
&M14 &3 &960 &[4, 2, 2] &[3840, 3008, 3840] &34.4 \\ 
&M15 &3 &960 &[8, 2, 8] &[4032, 4032, 3520] &36.1 \\ 
&M16 &3 &960 &[2, 2, 8] &[3584, 4032, 4032] &36.2 \\ 
&M17 &3 &960 &[2, 2, 8] &[4032, 4032, 3840] &36.7 \\ 
&M18 &3 &960 &[8, 4, 8] &[4032, 4032, 4032] &37.1 \\ 
&M19 &4 &896 &[4, 4, 8, 8] &[4032, 3456, 3328, 3392] &41.6 \\ 
&M20 &4 &960 &[4, 2, 8, 8] &[3840, 3008, 3328, 3584] &44.9 \\ 
&M21 &4 &960 &[2, 2, 8, 8] &[4032, 3968, 3904, 3840] &48.7 \\ 
&M22 &4 &960 &[4, 2, 4, 4] &[3840, 4032, 3904, 4032] &48.8 \\ 
&M23 &5 &960 &[4, 2, 4, 4, 8] &[3840, 3008, 3392, 2496, 4032] &55.3 \\ 
&M24 &5 &960 &[4, 2, 8, 8, 8] &[3840, 3008, 3840, 3328, 3968] &57.6 \\ 
&M25 &5 &960 &[2, 2, 4, 4, 4] &[3968, 4032, 3328, 4032, 2752] &57.9 \\ 
&M26 &6 &896 &[8, 4, 8, 4, 8, 8] &[3328, 2112, 3392, 3904, 3328, 3264] &58.8 \\ 
&M27 &5 &960 &[8, 2, 8, 8, 4] &[4032, 3008, 3840, 3904, 3968] &59.1 \\ 
&M28 &5 &960 &[2, 4, 2, 2, 8] &[3968, 3968, 3840, 4032, 3904] &60.9 \\ 
&M29 &6 &896 &[2, 2, 4, 4, 2, 2] &[3840, 3968, 3840, 3328, 3904, 3904] &65.0 \\ 
&M30 &6 &960 &[4, 8, 8, 4, 8, 4] &[3072, 3584, 3392, 3840, 3328, 3712] &67.9 \\ 
&M31 &6 &960 &[4, 8, 8, 8, 4, 4] &[3840, 3584, 3392, 3328, 3968, 3776] &69.7 \\ 
&M32 &6 &960 &[4, 8, 8, 8, 8, 2] &[3840, 3840, 3392, 3840, 3328, 3712] &69.9 \\ 
&M33 &6 &960 &[4, 2, 2, 4, 2, 8] &[3840, 3008, 3840, 3904, 4032, 3392] &70.0 \\ 
&M34 &6 &960 &[2, 4, 8, 8, 4, 2] &[3840, 3968, 3840, 3328, 4032, 3776] &71.5 \\ 
&M35 &7 &960 &[4, 8, 8, 8, 8, 2, 8] &[3840, 3968, 3840, 3328, 3968, 3328, 4032] &82.8 \\ 
&M36 &8 &960 &[4, 2, 8, 8, 8, 4, 8, 8] &[3840, 3968, 3840, 3328, 3072, 3328, 4032, 3072] &91.6 \\ 
&M37 &10 &896 &[8, 4, 8, 8, 8, 2, 8, 2, 4, 8] &[4032, 3008, 3840, 2560, 3904, 3904, 3072, 3264, 2368, 2496] &98.4 \\ 
&M38 &12 &832 &[2, 4, 8, 8, 8, 8, 8, 8, 8, 8, 4, 2] &[3840, 2816, 2112, 3584, 3648, 2432, 2304, 3008, 2880, 1664, 2432, 3776] &99.0 \\ 
&M39 &9 &960 &[8, 8, 8, 4, 4, 8, 8, 4, 2] &[2752, 3456, 2880, 3904, 2752, 3904, 4032, 3264, 3136] &99.3 \\ 
&M40 &10 &896 &[8, 8, 8, 2, 8, 2, 2, 2, 8, 2] &[3840, 3072, 3840, 2560, 3648, 3328, 3840, 3008, 2880, 3328] &100.0 \\ 
\hline
\end{tabular}}
\end{table}

\begin{table}[h]
\caption{Detailed architectural parameters for all models in Figure~\ref{fig:nas_results} with GPT-2 backbone.}\label{tab:archs_gpt}
\centering
\resizebox{\textwidth}{!}{
\begin{tabular}{lcccccc} \hline
\multicolumn{2}{c}{}
& n\textsubscript{layer} 
& d\textsubscript{model} 
& n\textsubscript{head} 
& d\textsubscript{inner}
& DecoderParams (M) \\ \hline 
\multicolumn{2}{l}{baseline} & $\in$[1,16] & 1024 & 12 & 3072 & - \\ \hline
\multirow{39}{*}{\rotatebox{90}{TITAN Xp}}&M1 &3 &256 &[2, 2, 2] &[3072, 3776, 3904] &6.3 \\ 
&M2 &2 &448 &[2, 2] &[3456, 3776] &8.1 \\ 
&M3 &2 &448 &[2, 4] &[4032, 3904] &8.7 \\ 
&M4 &3 &384 &[2, 2, 2] &[3072, 2176, 4032] &8.9 \\ 
&M5 &2 &576 &[2, 2] &[3456, 3584] &10.8 \\ 
&M6 &4 &448 &[2, 2, 2, 2] &[4032, 3904, 1920, 3072] &14.8 \\ 
&M7 &4 &512 &[2, 2, 4, 2] &[3904, 3136, 1280, 2624] &15.4 \\ 
&M8 &2 &832 &[8, 2] &[3456, 3584] &17.3 \\ 
&M9 &2 &960 &[2, 8] &[3456, 3648] &21.0 \\ 
&M10 &2 &960 &[2, 2] &[3968, 3584] &21.9 \\ 
&M11 &5 &640 &[2, 2, 2, 2, 2] &[4032, 2560, 2176, 2304, 3136] &26.4 \\ 
&M12 &3 &832 &[2, 8, 4] &[3840, 3840, 3776] &27.4 \\ 
&M13 &5 &704 &[2, 2, 2, 4, 4] &[2368, 3648, 1856, 3712, 3200] &30.8 \\ 
&M14 &3 &960 &[2, 2, 2] &[3584, 3648, 4032] &32.7 \\ 
&M15 &3 &960 &[2, 2, 2] &[3904, 3520, 4032] &33.1 \\ 
&M16 &6 &640 &[2, 2, 2, 2, 2, 2] &[2624, 2560, 2880, 3776, 3648, 3840] &34.6 \\ 
&M17 &4 &896 &[2, 2, 4, 2] &[4032, 3712, 3328, 3072] &38.2 \\ 
&M18 &5 &832 &[2, 2, 2, 4, 4] &[3392, 3648, 2880, 3712, 3200] &41.9 \\ 
&M19 &4 &960 &[2, 2, 4, 2] &[3904, 3136, 3328, 3776] &42.0 \\ 
&M20 &4 &960 &[8, 8, 2, 4] &[3904, 3712, 4032, 3776] &44.4 \\ 
&M21 &6 &832 &[2, 2, 4, 2, 2, 2] &[3904, 3456, 4032, 1792, 3072, 2496] &47.9 \\ 
&M22 &5 &896 &[4, 2, 2, 2, 4] &[3968, 3200, 3840, 3328, 3648] &48.3 \\ 
&M23 &5 &960 &[2, 2, 2, 2, 2] &[3904, 3264, 3328, 3776, 3392] &52.4 \\ 
&M24 &5 &960 &[2, 2, 4, 2, 2] &[3584, 3456, 3776, 2944, 4032] &52.7 \\ 
&M25 &5 &960 &[2, 8, 2, 4, 2] &[3904, 3648, 4032, 3776, 3968] &55.6 \\ 
&M26 &6 &960 &[8, 8, 2, 2, 2, 2] &[3904, 2560, 2880, 3776, 2240, 3840] &59.1 \\ 
&M27 &6 &960 &[2, 2, 2, 4, 2, 2] &[2496, 3456, 3328, 3904, 3968, 2944] &60.8 \\ 
&M28 &6 &960 &[4, 2, 4, 4, 2, 8] &[4032, 3456, 3328, 3776, 4032, 2752] &63.2 \\ 
&M29 &6 &960 &[2, 2, 2, 4, 4, 4] &[3968, 3648, 3840, 3776, 3584, 2624] &63.4 \\ 
&M30 &7 &960 &[2, 2, 2, 4, 2, 4, 2] &[3904, 2368, 4032, 3008, 3520, 2944, 2496] &68.7 \\ 
&M31 &7 &960 &[2, 2, 4, 2, 2, 2, 4] &[3072, 3648, 3520, 3584, 3136, 1984, 3584] &69.1 \\ 
&M32 &7 &960 &[4, 2, 2, 2, 8, 2, 2] &[3712, 3648, 3584, 3520, 2752, 3008, 3392] &71.2 \\ 
&M33 &8 &960 &[2, 4, 4, 2, 2, 2, 2, 2] &[3904, 2816, 3072, 1920, 3328, 3456, 2304, 2368] &74.1 \\ 
&M34 &8 &960 &[2, 2, 2, 4, 2, 2, 8, 2] &[3520, 2368, 4032, 1792, 3200, 3776, 3200, 3648] &78.6 \\ 
&M35 &8 &960 &[4, 2, 4, 4, 8, 8, 4, 2] &[3520, 3712, 3328, 3776, 3200, 2752, 3200, 2112] &78.7 \\ 
&M36 &8 &960 &[8, 4, 2, 8, 2, 2, 2, 2] &[3520, 3840, 3328, 3776, 3200, 3776, 3968, 3648] &85.4 \\ 
&M37 &10 &960 &[2, 8, 2, 4, 2, 2, 4, 2, 8, 8] &[3648, 2560, 3776, 1792, 3968, 2752, 3200, 2368, 4032, 2368] &95.5 \\ 
&M38 &10 &960 &[2, 4, 2, 2, 4, 2, 4, 2, 4, 8] &[3840, 2240, 3328, 3776, 3648, 3200, 2944, 2368, 3968, 2880] &98.8 \\ 
&M39 &10 &960 &[2, 4, 2, 2, 2, 2, 4, 2, 4, 8] &[3840, 2240, 3328, 3776, 3200, 3200, 3968, 2368, 3968, 2816] &99.8 \\ 
\hline
\end{tabular}}
\end{table}

\begin{table}[h]
\caption{Detailed architectural parameters for all models in Figure~\ref{fig:nas_results} with GPT-2 backbone.}\label{tab:archs_gpt2}
\resizebox{\textwidth}{!}{
\begin{tabular}{lcccccc} \hline
\multicolumn{2}{c}{}
& n\textsubscript{layer} 
& d\textsubscript{model} 
& n\textsubscript{head} 
& d\textsubscript{inner}
& DecoderParams (M) \\ \hline 
\multicolumn{2}{l}{baseline} & $\in$[1,16] & 1024 & 12 & 3072 & - \\ \hline
\multirow{36}{*}{\rotatebox{90}{ARM}}&M1 &2 &512 &[2, 2] &[1920, 1920] &6.0 \\ 
&M2 &3 &320 &[8, 2, 4] &[1920, 1920, 3712] &6.1 \\ 
&M3 &2 &576 &[2, 2] &[1344, 3200] &7.9 \\ 
&M4 &3 &384 &[2, 8, 2] &[3840, 2368, 3328] &9.1 \\ 
&M5 &5 &384 &[4, 4, 2, 4, 4] &[2880, 1920, 960, 2496, 1280] &10.3 \\ 
&M6 &2 &768 &[2, 2] &[1600, 2240] &10.6 \\ 
&M7 &5 &320 &[4, 2, 2, 4, 2] &[1344, 2240, 3776, 3008, 3648] &11.0 \\ 
&M8 &3 &768 &[2, 2, 4] &[1856, 1792, 1920] &15.7 \\ 
&M9 &3 &704 &[2, 2, 2] &[3136, 2112, 1920] &16.1 \\ 
&M10 &2 &960 &[4, 2] &[3584, 2304] &18.7 \\ 
&M11 &6 &448 &[4, 4, 2, 2, 4, 2] &[3072, 2112, 4032, 2688, 1600, 3072] &19.7 \\ 
&M12 &3 &960 &[4, 4, 2] &[2368, 2560, 2048] &24.5 \\ 
&M13 &4 &704 &[4, 8, 4, 2] &[3008, 3776, 2560, 3648] &26.3 \\ 
&M14 &5 &704 &[4, 2, 4, 2, 8] &[3584, 3136, 3776, 3072, 1856] &31.7 \\ 
&M15 &3 &960 &[2, 2, 2] &[3392, 3648, 3840] &32.0 \\ 
&M16 &4 &960 &[4, 2, 8, 2] &[2048, 3328, 1984, 1856] &32.5 \\ 
&M17 &7 &704 &[2, 4, 4, 4, 8, 2, 2] &[3008, 2560, 1920, 1856, 2112, 1728, 3136] &36.9 \\ 
&M18 &4 &960 &[2, 2, 4, 8] &[3392, 3456, 2432, 2304] &37.0 \\ 
&M19 &5 &832 &[4, 4, 4, 4, 4] &[3840, 1920, 4032, 3072, 3968] &41.9 \\ 
&M20 &5 &960 &[8, 4, 2, 2, 4] &[2560, 2048, 3648, 1728, 2304] &42.1 \\ 
&M21 &5 &960 &[4, 4, 2, 2, 2] &[3072, 2240, 1984, 2176, 3520] &43.4 \\ 
&M22 &5 &960 &[2, 4, 4, 4, 2] &[2496, 3648, 3328, 3392, 2112] &47.2 \\ 
&M23 &6 &832 &[4, 2, 4, 4, 2, 4] &[2496, 3200, 1664, 3904, 3520, 3840] &47.7 \\ 
&M24 &6 &960 &[8, 2, 2, 2, 8, 4] &[2304, 3328, 3456, 1856, 1792, 2112] &50.7 \\ 
&M25 &5 &960 &[4, 8, 2, 4, 4] &[3264, 2688, 4032, 3968, 3712] &52.4 \\ 
&M26 &6 &960 &[2, 4, 4, 2, 2, 2] &[3008, 2624, 4032, 2688, 3520, 2624] &57.7 \\ 
&M27 &6 &960 &[2, 4, 4, 2, 8, 2] &[2304, 3648, 3328, 3648, 3904, 1728] &57.8 \\ 
&M28 &6 &960 &[4, 4, 2, 4, 2, 2] &[3072, 2368, 4032, 4032, 3776, 3264] &61.6 \\ 
&M29 &7 &960 &[2, 2, 2, 8, 4, 8, 4] &[3008, 2304, 1920, 1984, 3520, 2816, 3712] &62.9 \\ 
&M30 &7 &960 &[2, 4, 4, 4, 4, 2, 2] &[3200, 4032, 2048, 2624, 2112, 2752, 2880] &63.6 \\ 
&M31 &7 &960 &[2, 4, 4, 4, 4, 2, 4] &[3584, 3648, 3328, 3392, 3200, 1984, 3200] &68.8 \\ 
&M32 &7 &960 &[2, 4, 8, 8, 2, 2, 8] &[3008, 3648, 3584, 3648, 3008, 1728, 3712] &68.8 \\ 
&M33 &7 &960 &[4, 4, 2, 4, 4, 8, 4] &[3584, 3840, 3328, 3392, 3136, 2944, 2496] &69.5 \\ 
&M34 &8 &960 &[8, 2, 2, 8, 2, 2, 8, 2] &[3008, 3648, 1792, 1984, 3008, 2816, 3712, 3520] &74.7 \\ 
&M35 &8 &960 &[2, 2, 2, 2, 8, 4, 4, 2] &[3008, 2304, 1792, 3008, 3520, 2880, 3712, 3456] &75.1 \\ 
&M36 &8 &960 &[2, 2, 2, 2, 2, 2, 4, 8] &[3008, 1792, 3840, 3392, 3520, 3136, 3712, 3520] &79.4 \\ 
&M37 &9 &960 &[2, 2, 4, 4, 8, 8, 4, 2, 4] &[1664, 1792, 2240, 3904, 3648, 3264, 2176, 3712, 1856] &79.9 \\ 
&M38 &11 &832 &[8, 4, 2, 4, 4, 2, 8, 4, 4, 8, 8] &[3072, 2368, 4032, 3968, 1664, 3968, 2176, 2624, 3840, 2176, 2112] &83.8 \\ 
&M39 &9 &960 &[4, 2, 4, 8, 2, 2, 4, 2, 4] &[2496, 3648, 3328, 3392, 3648, 1728, 2880, 3520, 2368] &85.1 \\ 
&M40 &9 &960 &[4, 2, 4, 8, 4, 2, 4, 2, 4] &[3072, 2816, 4032, 2560, 3648, 1728, 3840, 3264, 3456] &87.8 \\ 
&M41 &10 &960 &[8, 2, 4, 4, 2, 2, 4, 8, 2, 4] &[3648, 1792, 2432, 1856, 3392, 2304, 3776, 2944, 3136, 3904] &93.0 \\ 
&M42 &10 &960 &[8, 2, 2, 4, 2, 2, 2, 4, 2, 2] &[3264, 2048, 3520, 3904, 3840, 3840, 2624, 3072, 3776, 2304] &98.8 \\ 
&M43 &12 &896 &[4, 4, 4, 2, 4, 2, 4, 8, 8, 2, 4, 2] &[2048, 3136, 4032, 1792, 3584, 1728, 3136, 3008, 2560, 3200, 3648, 1728] &98.9 \\ 
&M44 &10 &960 &[4, 2, 8, 4, 2, 8, 4, 4, 4, 2] &[3584, 3968, 3328, 3904, 2368, 2112, 3904, 3520, 3328, 2688] &99.8 \\ 
&M45 &10 &960 &[8, 2, 4, 4, 4, 4, 4, 2, 2, 8] &[2688, 3200, 3840, 3392, 3520, 3136, 3392, 3520, 2880, 3200] &99.9 \\ 
\hline
\multirow{31}{*}{\rotatebox{90}{Corei7}}&M1 &2 &384 &[2, 2] &[3840, 2432] &6.0 \\ 
&M2 &3 &320 &[2, 2, 2] &[2176, 3072, 2496] &6.2 \\ 
&M3 &2 &512 &[2, 2] &[1408, 2624] &6.2 \\ 
&M4 &3 &384 &[2, 2, 2] &[3264, 3456, 3584] &9.7 \\ 
&M5 &2 &576 &[2, 2] &[3136, 3648] &10.5 \\ 
&M6 &3 &448 &[2, 2, 2] &[4032, 3648, 4032] &12.9 \\ 
&M7 &4 &448 &[2, 2, 4, 4] &[3072, 3648, 4032, 1792] &14.5 \\ 
&M8 &2 &768 &[2, 2] &[3968, 3328] &15.9 \\ 
&M9 &4 &576 &[2, 2, 2, 2] &[3072, 2752, 3456, 3136] &19.6 \\ 
&M10 &2 &960 &[2, 2] &[3840, 3264] &21.0 \\ 
&M11 &4 &640 &[2, 2, 2, 2] &[2176, 3648, 3584, 1920] &21.1 \\ 
&M12 &3 &960 &[2, 2, 2] &[2176, 3264, 2432] &26.2 \\ 
&M13 &4 &768 &[2, 2, 2, 2] &[3584, 2112, 3392, 1920] &26.4 \\ 
&M14 &4 &768 &[2, 2, 2, 2] &[3584, 2560, 3776, 1536] &27.1 \\ 
&M15 &4 &832 &[2, 2, 2, 2] &[3904, 1984, 3392, 3136] &31.8 \\ 
&M16 &3 &960 &[2, 2, 2] &[3968, 4032, 2880] &32.0 \\ 
&M17 &5 &768 &[2, 2, 4, 2, 2] &[3648, 3072, 3392, 1984, 2944] &34.9 \\ 
&M18 &4 &960 &[2, 2, 2, 2] &[3136, 1984, 3392, 2944] &36.8 \\ 
&M19 &4 &960 &[2, 2, 2, 4] &[3968, 3456, 3584, 3136] &42.0 \\ 
&M20 &6 &768 &[4, 2, 2, 4, 2, 4] &[3584, 2112, 3456, 3136, 3840, 2560] &42.9 \\ 
&M21 &7 &768 &[2, 4, 2, 4, 4, 4, 2] &[2624, 1984, 2496, 3968, 2880, 2112, 4032] &47.5 \\ 
&M22 &5 &960 &[2, 2, 4, 2, 4] &[2176, 3264, 3392, 3008, 3328] &47.6 \\ 
&M23 &6 &960 &[4, 4, 2, 4, 2, 2] &[2048, 2624, 3520, 1984, 2880, 2624] &52.3 \\ 
&M24 &6 &960 &[2, 4, 4, 4, 2, 2] &[1792, 3456, 2752, 2240, 1664, 3840] &52.4 \\ 
&M25 &6 &960 &[4, 2, 2, 2, 4, 4] &[2176, 1664, 3648, 3136, 3968, 3904] &57.7 \\ 
&M26 &7 &960 &[2, 2, 4, 4, 2, 2, 8] &[2816, 1792, 3968, 1728, 1664, 3328, 2944] &60.9 \\ 
&M27 &7 &896 &[2, 2, 4, 2, 2, 2, 2] &[3904, 3264, 3328, 3968, 1728, 2624, 4032] &63.5 \\ 
&M28 &7 &960 &[4, 2, 4, 2, 2, 2, 2] &[3584, 2560, 1792, 1920, 3968, 2112, 3968] &64.1 \\ 
&M29 &8 &960 &[2, 2, 2, 4, 2, 2, 2, 4] &[3328, 2432, 2624, 2752, 1664, 2240, 2304, 2816] &68.3 \\ 
&M30 &7 &960 &[4, 2, 4, 2, 2, 2, 2] &[3904, 2304, 2368, 3584, 3264, 2880, 3904] &68.5 \\ 
&M31 &8 &960 &[4, 2, 4, 2, 2, 4, 2, 4] &[2560, 3648, 2624, 2112, 3328, 2112, 1792, 3328] &70.9 \\ 
&M32 &8 &960 &[4, 4, 4, 2, 2, 4, 2, 4] &[2560, 2304, 2624, 4032, 2688, 2624, 3840, 2816] &74.7 \\ 
&M33 &9 &960 &[2, 4, 2, 4, 2, 4, 2, 2, 4] &[3072, 3264, 2944, 1984, 2880, 3520, 2112, 2624, 1728] &79.6 \\ 
&M34 &10 &896 &[2, 2, 4, 2, 2, 2, 2, 2, 4, 2] &[2816, 3264, 3584, 1792, 3136, 3584, 2240, 2240, 1920, 2752] &81.2 \\ 
&M35 &9 &960 &[8, 2, 2, 2, 4, 4, 2, 4, 4] &[3904, 3648, 2432, 3136, 3264, 2816, 2240, 3072, 3840] &87.7 \\ 
&M36 &10 &960 &[4, 4, 2, 2, 4, 4, 2, 4, 4, 2] &[2176, 3264, 2752, 3136, 3968, 3520, 3776, 3328, 1728, 2496] &94.9 \\ 
&M37 &10 &960 &[4, 2, 4, 2, 2, 2, 2, 4, 2, 2] &[3904, 2112, 2496, 3968, 3968, 2624, 3904, 2304, 3200, 3840] &99.0 \\ 
&M38 &11 &960 &[4, 2, 2, 4, 2, 4, 2, 2, 4, 4, 4] &[2176, 4032, 3264, 3840, 2688, 1984, 1728, 2944, 1920, 2368, 3840] &99.8 \\
\hline
\end{tabular}}
\end{table}

%% file: 4_2_zero-cost-NAS.tex
\section{How Good is the Decoder Parameters Proxy for Pareto-frontier Search?}\label{sec:nas}
In this Section, we validate whether the decoder parameter count proxy actually helps find Pareto-frontier models which are close to the ground truth Pareto front.
We first fully train all $1200$ architectures sampled from the Transformer-XL backbone during the evolutionary search (\ref{alg:algorithm}). Using the validation perplexity obtained after full training, we rank all sampled architectures and extract the ground truth Pareto-frontier of perplexity versus latency. We train the models on the WikiText-103 dataset and benchmark Intel Xeon E5-2690 CPU as our target hardware platform for latency measurement in this experiment. 

\begin{SCfigure}[50][h]
    \centering
    \includegraphics[width=0.4\columnwidth]{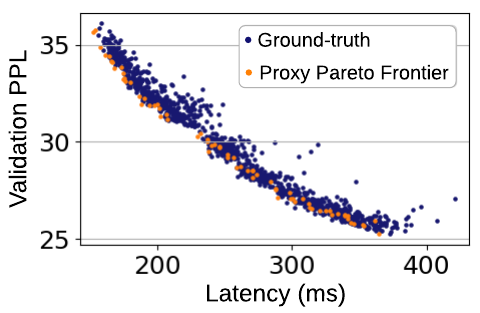}
    \vspace{-0.2cm}
    \caption{Perplexity versus latency Pareto obtained from full training of $1200$ architectures sampled during NAS on Transformer-XL backbone. Orange points are the Pareto-frontier extracted using decoder parameter count proxy, which lies close to the actual Pareto-frontier. Decoder parameter count holds an SRC of $0.98$ with the ground truth perplexity after full training.}
    \label{fig:nas}
\vspace{-0.2cm}
\end{SCfigure}

Figure~\ref{fig:nas} represents a scatter plot of the validation perplexity (after full training) versus latency for all sampled architectures during the search. The ground truth Pareto-frontier, by definition, is the lower convex hull of the dark navy dots, corresponding to models with the lowest validation perplexity for any given latency constraint. We mark the Pareto-frontier points found by the training-free proxy with orange color. As shown, the architectures that were selected as the Pareto-frontier by the proxy method are either on or very close to the ground truth Pareto-frontier. 


We define the mean average perplexity difference as a metric to evaluate the distance ($d_{avg}$) between the proxy and ground truth Pareto-frontier:
\addtolength{\abovedisplayskip}{-0.7ex}
\addtolength{\abovedisplayshortskip}{-0.7ex}
\addtolength{\belowdisplayskip}{-0.7ex}
\addtolength{\belowdisplayshortskip}{-0.7ex}
\begin{equation}
    d_{avg} = \frac{1}{N}\sum_{i=1}^N \frac{|p_i - p_{gt, i}|}{p_{gt, i}}  
\end{equation}
Here, $p_i$ denotes the $i$-th point on the proxy Pareto front and $p_{gt, i}$ is the closest point, in terms of latency, to $p_i$ on the ground truth Pareto front. The mean average perplexity difference for Figure~\ref{fig:nas} is $d_{avg}=0.6\%$. This small difference validates the effectiveness of our zero-cost proxy in correctly ranking the sampled architectures and estimating the true Pareto-frontier. In addition to the small distance between the prxoy-estimated Pareto-frontier and the ground truth, our zero-cost proxy holds a high SRC of $0.98$ over the entire Pareto, i.e., all $1200$ sampled architectures. 

\comment{
We provide the ranking performance for different evaluation techniques along with their cost in Table~\ref{tab:nas_cost}. As seen, our proposed evaluation proxy, i.e., decoder parameter count, provides the highest SRC with the ground truth ranking of the sampled architectures during NAS, while removing the extremely high overhead of training. 

\begin{SCtable}
\centering
\resizebox{0.5\columnwidth}{!}{
\begin{tabular}{lcccc}
\hline
& \begin{tabular}[c]{@{}c@{}}Train\\ Iter\end{tabular}
& \begin{tabular}[c]{@{}c@{}}GPU\\ Hours\end{tabular}
& \begin{tabular}[c]{@{}c@{}}$CO_2$e\\ (lbs)\end{tabular}
& SRC  
\\ \hline \hline
Full Training                     
& 40,000        
& 19,024    
& 5433                                                
& 1.0             
\\ \hline
\multirow{2}{*}{Partial Training} 
& 500           
& 231       
& 66                                                   
& 0.92            
\\ \cline{2-5}
& 5,000         
& 2690     
& 768                                                  
& 0.96            
\\ \hline
\# Decoder Params                 
& \textbf{0}             
& \textbf{0}         
& $\sim$\textbf{0}                                              
& \textbf{0.98}           
\\ \hline
\end{tabular}}
\caption{Comparison between training-based and the proposed training-free evaluation proxy on a real NAS benchmark. The SRC is computed on the entire population set of $1200$ models visited during the search. The training time is reported on the NVIDIA V100 GPU.}\label{tab:nas_cost}
\end{SCtable}
}

\begin{SCfigure}[50][h]
    \centering
    \includegraphics[width=0.35\columnwidth]{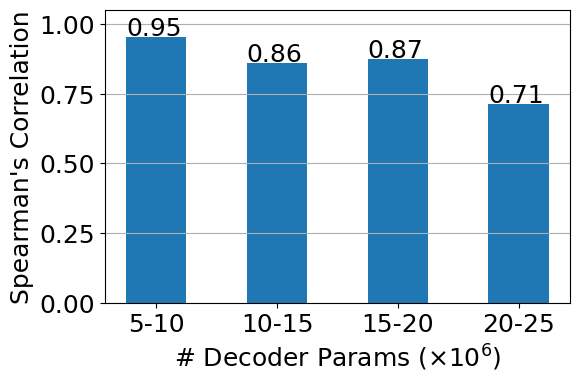}
    \vspace{-0.2cm}
    \caption{SRC between the decoder parameter count proxy and validation perplexity. Results are gathered on $1200$ models grouped into four bins based on their decoder parameter count. Our proxy performs well even when exploring within a small range of model sizes.}
    \label{fig:param_count_bins}
\vspace{-0.2cm}
\end{SCfigure}

We further study the decoder parameter proxy in scenarios where the range of model sizes provided for search is limited. We categorize the total $1200$ sampled architectures into different bins based on the decoder parameters. Figure~\ref{fig:param_count_bins} demonstrates the SRC between the decoder parameter count proxy and the validation perplexity after full training for different model sizes. The proposed proxy provides a highly accurate ranking of candidate architectures even when exploring a small range of model sizes. 

